\documentclass[letterpaper, 10 pt, journal, twoside]{IEEEtran}  

\IEEEoverridecommandlockouts                              

\usepackage{graphicx}
\usepackage{epsfig} 
\usepackage{mathptmx}
\usepackage{amsmath} 

\usepackage{amssymb}  
\usepackage{amsthm}
\usepackage{wasysym}
\usepackage[mathcal]{euscript}
\usepackage{bbm}
\usepackage{outlines}
\usepackage{color}
\usepackage{lipsum}
\usepackage[ruled,linesnumbered,vlined]{algorithm2e}
\usepackage{hyperref}
\usepackage{subfigure}
\usepackage{etoolbox}
\usepackage[table,xcdraw,dvipsnames]{xcolor}
\usepackage{multirow}

\usepackage{marginnote}
\newcounter{note}

\makeatletter
\patchcmd{\@makecaption}
  {\scshape}
  {}
  {}
  {}
\makeatother

\newtheorem{defn}{Definition}

\newtheorem{lem}[defn]{Lemma}
\newtheorem{prop}[defn]{Proposition}

\newtheorem{thm}[defn]{Theorem}

\providecommand{\R}{\ensuremath \mathbb{R}}
\providecommand{\N}{\ensuremath \mathbb{N}}
\newcommand{\zono}[1]{\mathcal{Z}\!\left(#1\right)}

\newcommand{\SO}{\mathsf{SO}}

\newcommand{\reachset}{\mathcal{R}}
\newcommand{\boxset}{\mathsf{box}}
\newcommand{\obsset}{\mathcal{O}}

\newcommand{\regtext}[1]{\mathrm{\textnormal{#1}}}

\newcommand{\defemph}[1]{\emph{#1}}
\newcommand{\ts}[1]{\textsuperscript{#1}}

\newcommand{\norm}[1]{\left\Vert#1\right\Vert}
\newcommand{\slicefunc}{\mathsf{slice}}

\newcommand{\minbbfunc}{\mathsf{minBoundingBox}}
\newcommand{\proj}{\mathsf{proj}}

\newcommand{\forwardoccupancy}{\mathsf{FO}}
\newcommand{\pow}[1]{\mathsf{pow}\left(#1\right)}
\newcommand{\diag}[1]{\mathsf{diag}\!\left(#1\right)}
\newcommand{\union}{\bigcup}

\newcommand{\trans}{^\top}
\newcommand{\inv}{^{-1}}
\newcommand{\adjustfunc}{\mathsf{adjust}}
\newcommand{\rewardfunc}{\mathsf{reward}}
\newcommand{\observationfunc}{\mathsf{observe}}
\newcommand{\rolloutfunc}{\mathsf{rollout}}
\newcommand{\sign}{\regtext{sign}}

\newcommand{\RL}{_{\regtext{RL}}}
\newcommand{\plan}{_{\regtext{plan}}}
\newcommand{\peak}{_{\regtext{des}}}

\newcommand{\des}{_{\regtext{des}}}
\newcommand{\slc}{_{\regtext{slc}}}
\newcommand{\frs}{_{\regtext{\tiny FRS}}}
\newcommand{\err}{_{\regtext{err}}}
\newcommand{\ini}{_\regtext{init}}

\newcommand{\desi}{\des}
\newcommand{\final}{_{\regtext{fin}}}
\newcommand{\obs}{_{\regtext{obs}}}

\newcommand{\submax}{_{\regtext{max}}}
\newcommand{\submin}{_{\regtext{min}}}
\newcommand{\safe}{_{\regtext{safe}}}

\newcommand{\sample}{_{\regtext{sample}}}
\newcommand{\hi}{}
\newcommand{\hiz}{_{0}}
\newcommand{\inione}{_{\regtext{init}}\arridx{1}}
\newcommand{\initwo}{_{\regtext{init}}\arridx{2}}
\newcommand{\desione}{_{\regtext{des}}\arridx{1}}
\newcommand{\desitwo}{_{\regtext{des}}\arridx{2}}

\newcommand{\track}{_{\regtext{trk}}}
\newcommand{\sense}{_{\regtext{sense}}}
\newcommand{\buf}{_{\regtext{extra}}}
\newcommand{\lng}{_{\regtext{long}}}
\newcommand{\lat}{_{\regtext{lat}}}
\newcommand{\cart}{_{\regtext{c}}}

\newcommand{\hifiddim}{{n_X}}
\newcommand{\Kdim}{{n_K}}
\newcommand{\ctrldim}{{n_U}}
\newcommand{\arbdim}{n} 

\newcommand{\plandim}{{n_P}}

\newcommand{\Kcenter}{c_K}
\newcommand{\Kdelta}{\Delta_K}

\newcommand{\numT}{{m_T}}
\newcommand{\numK}{{m_K}}
\newcommand{\numXhi}{{m_0}}

\newcommand{\idx}[1]{^{(#1)}} 
\newcommand{\arridx}[1]{^{[#1]}} 
\newcommand{\idxi}{\idx{i}}
\newcommand{\idxz}{\idx{0}}
\newcommand{\idxj}{\idx{j}}
\newcommand{\idxh}{\idx{h}}
\newcommand{\idxn}{\idx{n}}
\newcommand{\idxm}{\idx{m}}
\newcommand{\idxplan}{\idx{i,j}}
\newcommand{\idxerr}{\idx{i,j,h}}
\newcommand{\idxfrs}{\idx{i,j,h}}
\newcommand{\idxhp}{\idx{i,j,h,m}}

\newcommand{\gmp}{\gamma_{\regtext{p}}}
\newcommand{\gmd}{\gamma_{\regtext{d}}}

\newcommand{\PRS}{\reachset\plan}
\newcommand{\ERS}{\reachset\err}

\usepackage[
    style=ieee,
    doi=false,
    isbn=false,
    url=false,
    eprint=false,
    backend=biber,
    natbib=true
    ]{biblatex}
    
\bibliography{References}

\title{\LARGE \bf
Reachability-based Trajectory Safeguard (RTS): A Safe and Fast Reinforcement Learning Safety Layer for Continuous Control
}

\author{Yifei Simon Shao$^{1,2}$, Chao Chen$^{2}$, Shreyas Kousik$^3$, and Ram Vasudevan$^{1,2}$
\thanks{Manuscript received: October, 15, 2020; Revised February, 14, 2021; Accepted March, 1, 2021.}
\thanks{This paper was recommended for publication by Editor Clement Gosselin upon evaluation of the Associate Editor and Reviewers' comments.}
\thanks{This work is supported by the National Science Foundation Career Award \#1751093, the Ford Motor Company via the Ford-UM Alliance under award N022977, and the Office of Naval Research under Award Number N00014-18-1-2575.}
\thanks{$^{1}$Simon Shao and Ram Vasudevan are with the School of Mechanical Engineering, University of Michigan, Ann Arbor, MI. $\langle$\texttt{syifei, ramv}$\rangle$\texttt{@umich.edu}.}%
\thanks{$^{2}$Chao Chen is with the Robotics Institute, University of Michigan, Ann Arbor, MI. \texttt{joecc@umich.edu}.}%
\thanks{$^{3}$Shreyas Kousik is with the Department of Aeronautics and Astronautics, Stanford University, Stanford, CA. \texttt{skousik@stanford.edu}.}
\thanks{Digital Object Identifier (DOI): see top of this page.}
}

\begin{document}

\markboth{March, 2021}
{Shao \MakeLowercase{\textit{et al.}}: Reachability-based Trajectory Safeguard} 

\maketitle




\begin{abstract}
Reinforcement Learning (RL) algorithms have achieved remarkable performance in decision making and control tasks by reasoning about long-term, cumulative reward using trial and error.
However, during RL training, applying this trial-and-error approach to real-world robots operating in safety critical environment may lead to collisions. 
To address this challenge, this paper proposes a Reachability-based Trajectory Safeguard (RTS), which leverages reachability analysis to ensure safety during training and operation.
Given a known (but uncertain) model of a robot, RTS precomputes a Forward Reachable Set of the robot tracking a continuum of parameterized trajectories.
At runtime, the RL agent selects from this continuum in a receding-horizon way to control the robot; the FRS is used to identify if the agent's choice is safe or not, and to adjust unsafe choices.
The efficacy of this method is illustrated in static environments on three nonlinear robot models, including a 12-D quadrotor drone, in simulation and in comparison with state-of-the-art safe motion planning methods.
\end{abstract}


\section{Introduction}\label{sec:intro}

Reinforcement Learning (RL) is a powerful tool for automating decision making and control. 
For example, algorithms such as Soft Actor Critic (SAC) \cite{sac} and Twin Delayed DDPG (TD3) \cite{td3} have been successfully applied to operate robots in simulation.
The success is in part because RL attempts to maximize long term, cumulative reward.
However, RL suffers from a critical shortcoming: its inability to make safety guarantees during or after training.
Even when trained with a large penalty for collision, an RL agent may not be able to guarantee safety \cite{dalal2018safe}.

\begin{figure}[t]
\centering
\includegraphics[width=\columnwidth]{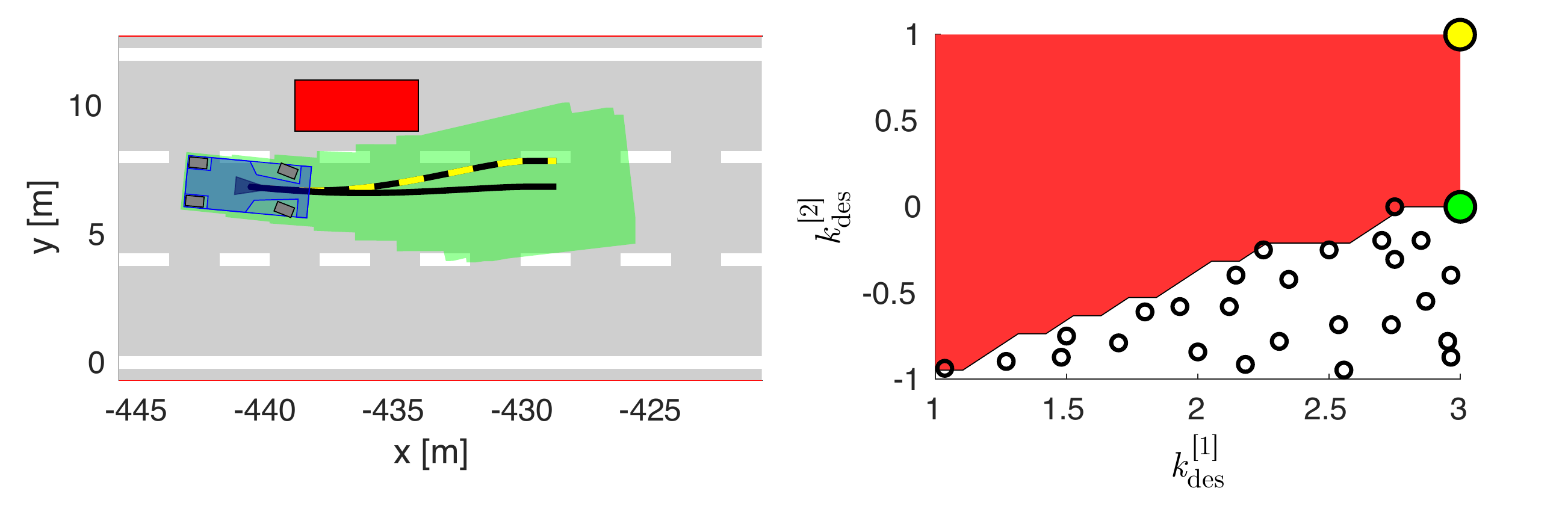}
\caption{An illustration of the proposed Reachability-based Trajectory Safeguard (RTS) enforcing safety during the training of a Reinforcement Learning (RL) policy.
The RL policy selects a trajectory parameter (yellow dot on right) which corresponds to a trajectory plan (yellow dashed line on left) for the car (blue box on left).
RTS efficiently identifies that this choice of trajectory parameter is unsafe with respect to the obstacle (red box on left), since it belongs to the unsafe trajectory parameter set (red set on right), computed via reachability analysis.
To replace the RL agent's plan with a new safe one, the parameter space is sampled (black circles), and the closest safe parameter (green dot on right) is selected.
Its trajectory (solid black line on left) and corresponding forward reachable set (green set on left) do not intersect the obstacle, guaranteeing safety despite tracking error.
By replacing the RL agent's plan, RTS enables learning from unsafe plans while the robot only executes safe plans.}
\label{fig:OnlineOps}
\vspace*{-0.3cm}
\end{figure}

\subsubsection*{Related Work}
A variety of techniques attempt to address the challenge of safe RL.
These can be broadly categorized as policy update, model update, and shield methods.

Policy update methods attempt to conservatively update a policy to maximize a reward function with constraint penalties in the reward function \cite{schulman2015trust,achiam2017constrained}.
Consequently, these methods typically only guarantee near constraint satisfaction.

Model update methods simultaneously learn a dynamic model and a stability or safety controller.
For example, one can learn to maximize a region of attraction by creating a Lyapunov function at each time step \cite{berkenkamp2017safe}.
Alternatively, a safe controller from a Control Barrier Function (CBF) \cite{cheng2019end} or the Hamilton-Jacobi-Bellman (HJB) equation \cite{fisac2018general,akametalu2018dissertation} can be applied to guide model learning. 
Finally, if one has a known control-invariant set, one can explore state space to improve a model until reaching a goal state \cite{lew2020safe}.
However, the challenges of constructing a Lyapunov function oe CBF, solving the HJB equation, or having a control-invariant set can restrict the complexity of models to which these techniques can be applied in real time.

Finally, shield methods adjust a policy at run-time using knowledge of the system's dynamics to ensure safety. 
For instance, one can linearize a robot model and apply quadratic programming to satisfy collision constraints \cite{dalal2018safe}. 
Similarly, one can attempt to find a parameterized trajectory that satisfies constraints while trying to maximize a desirable behavior \cite{shalev2016safe}.
Unfortunately, applying these methods in real time while making safety guarantees remains challenging. 
For real-time performance, one can adjust a policy by selecting from a finite, discrete (i.e., limited) set of actions \cite{TUM_safe_lane_change_set_based}. 

\subsubsection*{Present Work In Context}
We propose a safety layer using a trajectory-parameterized reachable set, computed offline, over a continuous action space, to ensure safe, real-time, online learning.
Unlike model update methods, we assume a known model, and compute bounded model uncertainty offline; and, instead of safe tracking of a given trajectory, we focus on receding-horizon trajectory design.
Unlike \cite{TUM_safe_lane_change_set_based}, we consider continuous instead of discrete actions, do not assume perfect trajectory-tracking, and consider more example robot models.
We also focus on reachability analysis of the ego robot, as opposed to other agents.

Our approach is motivated by recent applications of reachable sets for safe trajectory planning \cite{hess2014_parameterized_motion_primitives,pek2020online_verification,kousik2018bridging,vaskov2019towards,kousik2019safe}.
In particular, we extend Reachability-based Trajectory Design (RTD) \cite{kousik2018bridging}, which uses a simplified, parameterized model to make plans.
Offline, RTD upper bounds tracking error between the robot and planning model to find an over-approximating Forward Reachable Set (FRS) that describes the location of the robot for any plan.
At runtime, RTD uses the FRS to optimize over only safe plans.
However, RTD only optimizes over a short horizon, with a cost function generated by a high-level path planner.
Furthermore, RTD requires hand-tuning of the high-level planner to achieve frequent task completion.

Instead, this paper combines RTD's safety guarantees with RL's ability to maximize long term cumulative reward, eliminating high-level planner tuning.
We propose \defemph{Reachability based Trajectory Safeguard} (RTS), a safety layer for RL during both \textit{training} and \textit{runtime}.
We let the RL agent choose \textit{plan parameters} instead of control inputs.
This (1) lets us leverage reachability analysis to ensure plans are safe, and (2) reduces the dimension of the RL agent's action space, speeding up runtime operation.
Enforcing safety can also simplify reward tuning by removing the need for obstacle avoidance or collision penalties.

Note, we restrict our discussion to static environments to simplify exposition, since safety cannot be guaranteed when other agents may act maliciously.
However, like RTD \cite{vaskov2019not,vaskov2019towards,shreyas_dissertation}, this work can extend to dynamic environments by using predictions of other agents from, e.g., \cite{salzmann2020trajectron}.

\subsubsection*{Contributions}
The contributions of this work are three-fold.
First, we propose the RTS algorithm for safe, real-time RL training and deployment using a continuous action space.
Second, we reduce the conservatism of RTD's reachable sets with a novel tracking error representation.
Third, we demonstrate RTS in simulation on three nonlinear models/tasks: a 4-D cartpole swing up task on a limited track, a 5-D car on an obstacle course, and a 12-D quadrotor drone in a cluttered tunnel.
We compare against RTD, RTS with a discrete action space, and baseline RL.
RTS outperforms the other methods in terms of reward and tasks completed\footnote{Our code is available online at \url{www.github.com/roahmlab/reachability-based_trajectory_safeguard}.}.

\subsubsection*{Paper Organization}
In Sec. \ref{sec:robot_and_env}, we model the robot and the environment.
In Sec. \ref{sec:offline_reachability}, we compute reachable sets of the robot offline.
In Sec. \ref{sec:safe_RL}, we use the reachable sets online (during training) to ensure safety.
In Sec. \ref{sec:experiments}, we demonstrate the method.
Sec. \ref{sec:conclusion} provides concluding remarks.

\subsubsection*{Notation}
Points, vectors, and point- or vector-valued functions (resp. sets, arrays, and set- or array-valued functions) are in lowercase (resp. uppercase) italics.
The real numbers are $\R$; the natural numbers are $\N$.
For $n, m \in \N$, we let $\N_n = \{1,2,\cdots,n\} \subset \N$, and $\N_n + m = \{1+m,\cdots,n_m+m\}$.
The $n$-dimensional special orthogonal group is $\SO(n)$.
For a set $A$, its cardinality is $|A|$ and its power set is $\pow{A}$.
If $A$ is indexed by elements of $B$, we write $a\idx{b} \in A$ for $b \in B$.

Brackets denote concatenation when the size of vectors/matrices are important; e.g., if $v_1, v_2 \in \R^n$, then $[v_1, v_2] \in \R^{n\times 2}$ and $[v_1\trans, v_2\trans]\trans \in \R^{2n}$.
Otherwise, we write $(v_1,v_2) \in \R^{2n}$.
We use $\diag{v_1,v_2,\cdots,v_n}$ to place the elements of each input vector (in order) on the diagonal of a matrix with zeros elsewhere.
We denote an empty vector/matrix as $[\ ]$.

We denote a multi-index as $I = \{i_1 < i_2 < \cdots < i_p\} \subset \N_n$ with $i_p \leq n$, $p \leq n$, and $|I| = p$.
If $v \in \R^n$, $v\arridx{I} \in \R^p$ contains the elements of $v$ indexed by $I$.
Let $M \in \R^{n\times m}$ be a matrix; let $I_1 \subset \N_n$, and $I_2 \subset \N_m$, with $|I_1| = p \leq n$ and $|I_2| = q \leq m$.
Then $M\arridx{I_1,I_2}$ is a $p\times q$ sub-matrix of $M$ with the elements indexed by $I_1$ (for the rows) and $I_2$ (for the columns).
Similarly, $M\arridx{:,I_2}$ is the $n\times q$ sub-matrix of the columns of $M$ indexed by $I_2$.
If $M\idx{i}$ is the $i$\ts{th} matrix in a set, $[M\idx{i}]\arridx{j}$ is its $j$\ts{th} element.

Let $\boxset(c, l, R)\subset \R^\arbdim$ be a \defemph{rotated box}, with center $c \in \R^\arbdim$, edge lengths $l\in\R^\arbdim$, and rotation $R\in \SO(\arbdim)$:
\begin{align}\begin{split}
    \boxset(c,l,R) = c + R\cdot\Big([-l\arridx{1},l\arridx{1}]\times\cdots\times[-l\arridx{n},l\arridx{n}]\Big).
\end{split}\end{align}
If $R$ is an identity matrix, we say the box is \defemph{axis-aligned}.

\section{Robot and Environment}\label{sec:robot_and_env}

This section describes the robot and its surroundings.

\subsection{Modeling the Robot}\label{sec:robot_model}

\subsubsection{High Fidelity Model}\label{subsubsec:high-fid_model}

We express the robot's motion using a \defemph{high-fidelity model}, $f\hi: X\hi \times U \to \R^\hifiddim$, with state space $X\hi \subset \R^\hifiddim$, control input space $U \subset \R^\ctrldim$, and
\begin{align}
    \dot{x}\hi(t) = f\hi(x\hi(t),u),
\end{align}
where $t \in T = [0,t\final]$ is time in each planning iteration, $x\hi: T \to \R^\hifiddim$ is a trajectory of the high-fidelity model, and $u \in U$ is the control input.
We require that $f\hi$ is Lipschitz continuous, and $T$, $X\hi$, and $U$ are compact, so trajectories exist.
We assume $f\hi$ accurately describes the robot's dynamics.
One can extend the method to where this model only describes robot motion to within an error bound \cite[Thm. 39]{kousik2018bridging}.

Our goal is to represent rigid-body robots moving through space, so we require that the robot's state $x\hi \in X\hi$ includes the robot's position $p$ in a \defemph{position subspace} $P \subset \R^\plandim$ ($\plandim = 1, 2$, or $3$).
For example, $p$ can represent the robot's center of mass position in global coordinates.
We use a projection map $\proj_P: X\hi \to P$ to get the position from $x\hi \in X\hi$ as $p = \proj_P(x\hi)$.

\subsubsection{Planning Model}\label{subsec:planning_model}
We require frequent replanning for real-time operation; doing so is typically challenging with a high-fidelity model directly, so we use a simpler planning model and bound the resulting error.
Let $K \subset \R^\Kdim$ denote a compact space of \defemph{plan parameters} (detailed below).
The \defemph{planning model} is a map $p\plan: T\times K \to P$ that is smooth in $t$ and $k$, with $p\plan(0,k) = 0$ for all $k \in K$, and $\dot{p}\plan(t\final,k) = 0$ for all $k \in K$.
We refer to a single plan $p\plan(\cdot,k): T \to P$ as a \defemph{plan}.
Note, every plan begins at $t = 0$ without loss of generality (WLOG), and every plan is of duration $t\final$.
We use the smoothness of $p\plan$ to compute reachable sets in Section \ref{sec:offline_reachability}.
We fix $p\plan(0,k) = 0$ WLOG because we can translate/rotate any plan to the position/orientation of the high-fidelity model at the beginning of each planning iteration.
We fix $\dot{p}\plan(t\final,\cdot) = 0$ so all plans end with a braking \defemph{failsafe maneuver}.
If the robot fails to find a safe plan in a planning iteration, it can continue a previously-found safe plan, enabling persistent safety \cite[Sec. 5.1]{fraichard2004inevitable_coll_states}.

\subsubsection{The Plan Parameter Space}\label{subsubsec:K_defn}

We require that $K$ is an axis-aligned, box-shaped set.
Let $\Kcenter, \Kdelta \in \R^\Kdim$.
Then
\begin{align}
    K = \boxset(\Kcenter,\Kdelta,0).
\end{align}
We also break the parameter space into two subspaces, so that $K = K\ini \times K\desi$.
Denote $k = (k\ini,k\desi) \in K$.

The first subspace, $K\ini$, determines a plan's initial velocity, $\dot{p}\plan(0,k\ini)$; this ensures one can choose a plan that begins at the same velocity as the robot.
To this end, we define an \defemph{initial condition function}, $f\ini: X\hi \to K\ini$, for which $x\hi \mapsto k\ini$.
Suppose the robot is at a state $x\hi$ and applying a control input $u$.
We implement $f\ini$ by setting $\dot{p}\plan(0,k) = \proj_P(f\hi(x\hi,u))$ and solving for $k\ini$, where we have abused notation to let $\proj_P$ project the relevant coordinates.

The second subspace, $K\desi$, specifies positions or velocities reached by a plan during $(0,t\final] \subset T$.
So, instead of choosing control inputs, the RL agent chooses $k\desi$ in each receding-horizon planning iteration.
This design choice is an important feature of RTS+RL, because we can design a tracking controller (discussed below) to obtain stability guarantees and obey actuator limits, and let RL focus on decision making.
Note, different choices of $k\desi$ may be safe or unsafe, whereas $k\ini$ is determined by the robot's state at $t = 0$.

\subsubsection{Receding-horizon Timing}

We specify the rate of operation with a \defemph{planning time}, $t\plan \in (0,t\final)$.
In each planning iteration, if a new, safe plan is found before $t\plan$, the robot begins tracking it at $t\plan$.
Otherwise, the robot continues its previous plan.
We find the robot's initial condition $x\hiz$ for each plan by forward-integrating the high-fidelity model tracking the previous plan for duration $t\plan$.

\subsubsection{Tracking Controller and Error}
We use a \defemph{tracking controller}, $u\track: T\times X\hi\times K \to U$ to drive the high-fidelity model towards a plan, with \defemph{tracking error} $e: T \to \R^\plandim$ as
\begin{align}
    e(t;x\hiz,k) &= p(t;x\hiz,k) - p\plan(t,k)\ \regtext{and}\label{eq:tracking_error_defn}\\
    x\hi(t;k,x\hiz) &= \int_0^t f\hi(x\hi(s),u\track(t,x\hi(s),k)ds + x\hiz,\label{eq:hi_fid_model_traj_with_u_trk}
\end{align}
where $x\hiz \in X\hi$ such that $p(0;x\hiz,k) = 0$ and $\dot{p}\plan(0;k) = \dot{p}(0)$, with $p(\cdot) = \proj_P(x\hi(\cdot))$.
Note, $e$ is bounded because $f\hi$ is Lipschitz and $T$, $U$, and $K$ are compact.
We account for tracking error to ensure safety in Secs. \ref{sec:offline_reachability} and \ref{sec:safe_RL}.

\subsection{Modeling the Robot's Environment}
\subsubsection{Forward Occupancy}
The position coordinate $p \in P$ typically describes the motion of the robot's center of mass, but we require the robot's entire body to avoid obstacles.
So, we define the \defemph{forward occupancy map} $\forwardoccupancy: X\hi \to \pow{P}$, for which $\forwardoccupancy(x\hi) \subset P$ is the volume occupied by the robot at state $x\hi \in X\hi$.
Since we only consider rigid body robots in this work, we conflate the robot's workspace with $P$.
Note this work can extend to non-rigid robots such as multilink arms \cite{holmes2020reachable}.

\subsubsection{Safety and Obstacles}\label{subsubsec:safety_and_obstacles}

We define \defemph{safety} as collision avoidance.
Let an \defemph{obstacle} $O \subset P$ be a static region of workspace for which, if the robot is at a state $x\hi$ and $\forwardoccupancy(x\hi) \cap O \neq \emptyset$, the robot is \defemph{in collision}.
We assume each obstacle is a box, $O = \boxset(c,l,R) \subset P$, and static with respect to time; note that typical mapping and perception algorithms output rotated boxes \cite{yolov4,Thrun2003RoboticMA}.
We also assume that the robot need only avoid a finite number of obstacles for any plan.
We assume the robot can sense every obstacle within a finite distance $d\sense$ of its position, as in \cite[Thm. 39]{kousik2018bridging}.
Note the present work can extend to dynamic environments \cite{vaskov2019not,vaskov2019towards}.


\section{Offline Reachability Analysis}\label{sec:offline_reachability}

Offline, we compute a Planning Reachable Set (PRS), $\PRS$, of the planning model, then bound tracking error with an Error Reachable Set (ERS), $\ERS$.
Online, we use the PRS and ERS to build safety constraints (in the next section).

\subsection{Planning Reachable Set}\label{subsec:PRS}

We represent the PRS with zonotopes using an open-source toolbox called CORA \cite{althoff2015cora}.
A zonotope is a set
\begin{align}\label{eq:zono_defn}
    \zono{c,G} = \left\{y \in \R^n\ |\ y = c + G\beta,\ \beta \in [-1,1]^m\right\},
\end{align}
where $c \in \R^n$ is the zonotope's \defemph{center}, $G \subset \R^{n\times m}$ is a \defemph{generator matrix}, and $\beta$ is a \defemph{coefficient vector}.
The columns of $G$ are called \defemph{generators} of the zonotope.

\subsubsection{PRS Computation Setup}
We provide the toolbox \cite{althoff2015cora} with three inputs to compute the PRS.
First, we partition $T$ into $\numT \in \N$ intervals.
Let $\Delta_T = t\final/\numT$, $T\idx{1} = [0,\Delta_T]$, and $T\idxi = \big((i-1)\cdot\Delta_T, i\cdot\Delta_T\big]$.
Then, $T = \union_{i=1}^{\numT} T\idxi$.
Second, we provide an augmented state $z = (p,k)$ with $\dot{z} = (\dot{p}\plan, 0)$, to allow computing the PRS for all $k$.
Third, since for any $k$, a plan has $p\plan(0,k) = 0$, we provide an initial condition zonotope $Z\plan\idxz = \zono{c\idxz,G\idxz} \subset P\times K$, with center $c\idxz = (0_{\plandim\times 1},\Kcenter) \in \R^m$ and generator matrix $G\idxz = \diag{0_{\plandim\times 1},\Kdelta} \in \R^{m\times m}$, where $m = \plandim + \Kdim $.

\subsubsection{PRS Representation}
The PRS is represented as
\begin{align}\label{eq:prs}
    \PRS = \left\{Z\plan\idxi = \zono{c\plan\idxi,G\plan\idxi} \subset P\times K\ |\ i \in \N_{\numT}\right\},
\end{align}
which conservatively contains all plans and parameters: if $t \in T\idxi$ and $k \in K$, then $(p\plan(t,k),k) \in Z\plan\idxi$ \cite[Thm. 3.3]{althoff2010reachability}.

\subsubsection{Plan Parameter Partition}
In practice, the conservatism of the PRS zonotope representation is proportional to the size of $K\ini$.
So, we partition $K\ini$ into $\numK \in \N$ axis-aligned boxes $K\ini\idxj \subset K$ such that $K \subseteq \union_{j=1}^{\numK} (K\ini\idxj\times K\desi)$ 
and $K\ini\idxi \cap K\ini\idxj = \emptyset$ when $i \neq j$.
We compute \eqref{eq:prs} for each $K\idxj = K\ini\idxj\times K\desi$, and choose which PRS to use online (in each planning iteration) based on the robot's initial condition, by choosing $j$ such that $f\ini(x\hiz) \in K\ini\idxj$.

\subsection{Error Reachable Set}\label{subsec:ERS}

The ERS bounds tracking error as in \eqref{eq:tracking_error_defn}.
Novel to this work, we also use the ERS to bound the robot's forward occupancy in workspace, whereas \cite{kousik2019safe,shreyas_dissertation} bounded the occupancy in the PRS, which is more conservative in practice.

\subsubsection{Initial Condition Partition}
Notice that tracking error depends on the robot's initial condition $x\hiz \in X\hiz = \{x\hiz \in X\hi\ |\ \proj_P(x\hiz) = 0\}$.
Just as we partitioned $K\ini$ to reduce PRS conservatism, we partition $X\hiz$ to reduce ERS conservatism.
We choose $\numXhi \in \N$ axis-aligned boxes $X\hiz\idxh$ such that $X\hiz \subseteq \union_{h=1}^{\numXhi} X\hiz\idxh$,
and $X\hiz\idxi \cap X\hiz\idxj = \emptyset$ when $i \neq j$.

\subsubsection{Zonotope ERS}
With our partition of $X\hiz$, we represent the ERS as a collection of zonotopes
\begin{align}\label{eq:ers}
    \ERS = \{Z\err\idxfrs \subset P\ |\ (i,j,h) \in \N_{\numT}\times\N_{\numK}\times \N_{\numXhi}\}
\end{align}
for which, if $t \in T\idxi$, $k = (k\ini,k\desi) \in K\idxj$, $x\hiz \in X\hiz\idxh$, and $f\ini(x\hiz) = k\ini$, then
\begin{align}\label{eq:ers_conservatism}
    \forwardoccupancy(x\hi(t;k,x\hiz)) \subseteq \{p\plan(t,k)\} + Z\err\idxerr
\end{align}
with $x\hi$ as in \eqref{eq:hi_fid_model_traj_with_u_trk} and $+$ denoting set addition \cite[Lem. 6]{holmes2020reachable}.

\subsubsection{Computing the ERS via Sampling}\label{subsubsec:compute_ers_via_sampling}
It is challenging to compute \eqref{eq:ers} using reachability tools such as \cite{althoff2015cora}, because the high-dimensional, nonlinear tracking error results in excessive conservatism.
Instead, we use adversarial sampling, wherein we extend \cite[Alg. 3]{shreyas_dissertation} for forward occupancy.

\begin{figure}[t] \label{fig:err_bound}
\centering
\includegraphics[trim={3cm 0 6cm 0},clip,width=\linewidth]{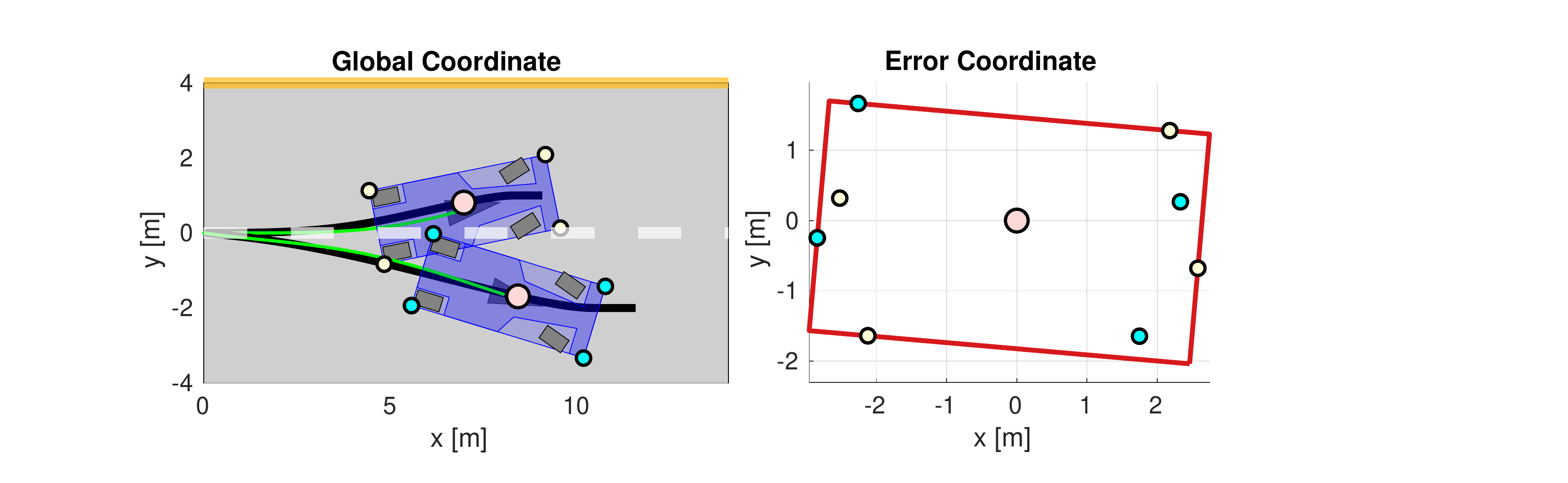}
\caption{An example of tracking error (as in \eqref{eq:tracking_error_defn}) plus the robot body at a sampled time, for two sampled plans (black on left).
The tracking error plus robot body (yellow and cyan dots) is shown with respect to the robot's center of mass (pink), and bounded with a rotated box (red) as in \eqref{eq:ers_zonotope_def}.}
\label{fig:ERS}
\vspace*{-0.35cm}
\end{figure}

Our goal is to conservatively estimate worst-case tracking error using a finite number of samples in $K\idxj\times X\hiz\idxh$ by leveraging the box structure of $K\idxj$ and $X\hiz\idxh$.
An axis-aligned box $B \subset \R^n$ can be expressed $[-l_1,l_1]\times\cdots\times[-l_n,l_n]$.
We call $\{-l_1,l_1\}\times\cdots\times\{-l_n,l_n\}$ the box's \defemph{corners}.
Let $C\idx{j,h}$ be the corners of  $K\idxj\times X\hiz\idxh$.
We sample each corner of each $C\idx{j,h}$.
Using all corners $(k,x\hiz) \in C\idx{j,h}$, we find the zonotope $Z\err\idxerr$ as follows.
First, if needed, we adjust $k\ini$ such that $f\ini(x\hiz) = k\ini$.
Then, we find $x\hi$ as in \eqref{eq:hi_fid_model_traj_with_u_trk} and
\begin{align}\label{eq:ers_volume}
    V\idxerr = \union_{(t,k,x\hiz) \in S} \forwardoccupancy(x\hi(t;k,x\hiz)) - \{p\plan(t,k)\},
\end{align}
with $S = T\idxi\times C\idx{j,h}$ and $-$ denoting set subtraction.
Note, in practice, we discretize $T\idxi$ to estimate this union, and numerically estimate $x\hi$ with a standard differential equation solver.
Finally, we compute each ERS zonotope as
\begin{align}\label{eq:ers_zonotope_def}
    Z\err\idxerr = \minbbfunc(V\idxerr),
\end{align}
where $\minbbfunc$ returns a minimum bounding box using \cite{bbox}.
We use rotated boxes to enable fast online planning.
Note, $\boxset(c,l,R) = \zono{c,R\,\diag{l}}$ by \eqref{eq:zono_defn}.
Fig. \ref{fig:ERS} shows an example of \eqref{eq:ers_zonotope_def}.
The proposed method reduces ERS conservatism compared to \cite{shreyas_dissertation}, which overapproximates all rotations of a robot's body with one zonotope.
However, our method is limited by exponential growth of the number of samples with the state dimension, and by finding the robot's high-fidelity model and ERS offline.

\subsubsection{Justifying Conservatism}
We now justify that our ERS sampling strategy can satisfy \eqref{eq:ers_conservatism}.
We improve upon \cite[Prop. 7.1]{shreyas_dissertation}, which justifies sampling the corners of each $K\ini\idxj$ and $X\hiz\idxh$, by justifying why we sample the corners of $K\idxj$.
To proceed, we assume our robot's actuators are modeled as double integrators, and that maximizing actuator tracking error maximizes robot tracking error.
Then, tracking error is proportional to commanded change in velocity:

\begin{prop}\label{prop:tracking_error_max_at_max_k_des}
Consider a 1-D actuator model with states $(p,\dot{p}) \in \R^2$ and $\ddot{p} = u \in \R$.
Let $p\plan(\cdot,k): T \to \R$ be a smooth plan.
Consider control gains $\gmp,\gmd \in \R$, and let
\begin{align}\label{eq:prop_k_desi_controller}
    u = \gmp\cdot(p - p\plan) + \gmd\cdot(\dot{p} - \dot{p}\plan),
\end{align}
Suppose $p(0) = p\plan(0,k)$ and $k = (k\ini,k\desi)$ with $k\ini$ fixed such that $\dot{p}(0,k) = \dot{p}\plan(0,k)$.
Suppose that, for all $t \in T$, $\ddot{p}\plan(t,k) = k\desi \in [k\submin,k\submax] \subset \R$.
Then the tracking error $|p(t) - p\plan(t)|$ is maximized when $k\desi \in \{k\submin,k\submax\}$.
\end{prop}
\begin{proof}
Consider the tracking error system
\begin{align}
    z(t,k) &= \begin{bmatrix}
            z_1(t,k) \\ z_2(t,k)
        \end{bmatrix} =
        \begin{bmatrix}
            p(t) - p\plan(t,k) \\ \dot{p}(t) - \dot{p}\plan(t,k)
        \end{bmatrix}.
\end{align}
Recalling that $\dot{k} = 0$ for any plan, we have
\begin{align}
    \dot{z}(t,k) &= \underset{A}{\underbrace{\begin{bmatrix}
            0 & 1 \\ \gmp & \gmd
        \end{bmatrix}}}z(t,k) +
        \begin{bmatrix}
            0 \\ \ddot{p}\plan(t,k)
        \end{bmatrix},\label{eq:prop_k_desi_proof_first_order_system}
\end{align}
for any fixed $k \in K$.
We can solve for $z$ to find
\begin{align}
    z(t,k) = -A\inv(e^{At} - I_{2\times 2})\begin{bmatrix}
            0 \\ k\desi
        \end{bmatrix},\label{eq:prop_k_desi_proof_particular_soln}
\end{align}
where $I_{2\times 2}$ is an identity matrix.
Notice that
\begin{align}
    A\inv = \begin{bmatrix}
            -\gmd/\gmp & 1/\gmp \\ 1 & 0
        \end{bmatrix}~\implies~e^{At} = \begin{bmatrix}
            a_1(t) & a_2(t) \\ a_3(t) & a_4(t)
        \end{bmatrix},
\end{align}
where we can choose $\gmp$ and $\gmd$ such that $a_2(t)$ and $a_4(t) \neq 0$.
Then, by expanding \eqref{eq:prop_k_desi_proof_particular_soln}, we have
\begin{align}
    z_1(t,k) = \frac{1 - \gmd a_2(t) + a_4(t)}{\gmp} k\desi,
\end{align}
completing the proof.
\end{proof}
\noindent Note that the planning model in Prop. \ref{prop:tracking_error_max_at_max_k_des} does not obey $\dot{p}\plan(t\final,k) = 0$.
This simplification is to illustrate the main idea: the tracking error is proportional to $k\desi$.
However, a similar result holds when $\ddot{p}\plan(t,k) \propto (k\desi - k\ini)t^2$ (as is true for the planning models in Sec. \ref{sec:experiments}) by applying integration by parts to solve \eqref{eq:prop_k_desi_proof_first_order_system}.
Furthermore, input saturation does not affect the result of Prop. \ref{prop:tracking_error_max_at_max_k_des}, because then $\ddot{p}$ would be constant for some duration.
So, one can maximize tracking error by choosing $k\des$ to maximize input saturation.

\section{Online Safe Reinforcement Learning}\label{sec:safe_RL}

This section describes online training and testing with RTS, wherein the robot only chooses safe plans while \textit{still learning from unsafe plans}.
The main result is in Thm. \ref{thm:safe_RL}.

To enforce safety, we combine the PRS (i.e., all plans) and ERS (i.e. tracking error) to build a Forward Reachable Set (FRS) containing the motion of the high-fidelity model tracking any plan.
If a subset of the FRS corresponding to a plan is collision free, then that plan is safe.

\subsection{Reachability-based Trajectory Safeguard (RTS)}\label{subsec:RTS}

Consider a single planning iteration.
Suppose the robot is generating a plan beginning from an initial condition $x\hiz \in X\hi$.
We present all computations from here on in the robot's local coordinate frame at $t = 0$, so $\proj_P(x\hiz) = 0$.
We build the FRS, use it to create safety constraints, then present an algorithm to ensure the robot only chooses safe plans.

Before we construct the FRS, we choose the PRS and ERS zonotopes from the partition of $K$ and $X\hiz$.
Pick $x\hiz \in X\hiz\idxh$ and $K\idxj$ such that if $k\ini = f\ini(x\hiz)$, then $k = (k\ini,k\desi) \in K\idxj$.
Let $i \in \N_{\numT}$ (i.e., the time interval $T\idxi$) be arbitrary.
For the rest of this section, we consider the zonotopes
\begin{align}
    Z\plan\idxplan &= \zono{c\plan\idxplan,G\plan\idxplan} \subset P\times K\ \regtext{and}\\
    Z\err\idxerr &= \zono{c\err\idxerr,G\err\idxerr} \subset P.
\end{align}

\subsubsection{FRS Construction}\label{subsec:frs_online}

We represent the FRS as zonotopes $Z\frs\idxfrs \subset P\times K$ built from the PRS and ERS zonotopes:
\begin{align}\label{eq:ERS_zono_plus_PRS_zono}
    Z\frs\idxfrs = \zono{c\plan\idxplan + \begin{bmatrix} c\err\idxerr \\ 0_{\Kdim\times 1} \end{bmatrix}, \left[G\plan\idxplan,\ \begin{bmatrix}
            G\err\idxerr \\ 0_{\Kdim \times \plandim}
        \end{bmatrix}\right]},
\end{align}
which follows from the zonotope Minkowski sum and Cartesian product \cite{guibas2003zonotopes}.
Note, the first $\plandim$ rows of $G\plan\idxplan$ correspond to all $\plandim$ rows of $G\err\idxerr$ (and similarly for the centers).

\subsubsection{Creating Safety Constraints}\label{subsec:creating_safety_cons}

Let $\{O\idxm\}_{m=1}^{n\obs}$ be the set of obstacles that the robot must avoid in the current planning iteration.
The robot must choose $k\desi$ such that, if $k = (k\ini,k\desi)$, then $\forwardoccupancy(x\hi(t;x\hiz,k))\cap O\idxm = \emptyset\  \forall\ m$.
To check this intersection, we introduce \defemph{slicing}.
Let $c \in \R^n, G \in \R^{n\times p}$.
Let $I \subseteq \N_p$ be a multi-index and $\beta \in [-1,1]^{|I|}$.
We define
\begin{align}\label{eq:slice_defn}
    \slicefunc(\zono{c,G},I,\beta) = \zono{c + G\arridx{:,I}\beta,\ G\arridx{:,\N_p\setminus I}}.
\end{align}
From \eqref{eq:zono_defn}, a sliced zonotope is a subset of the original zonotope, reducing the conservatism of using reachable sets for online planning.
We use slicing to identify unsafe plans:

\begin{lem}\label{lem:test_if_sliced_FRS_intersects_obs}
Consider the obstacle $O\idxm = \zono{c\obs\idxm,G\obs\idxm} \subset P$ and denote $Z\frs\idxfrs = \zono{c\frs\idxfrs, G\frs\idxfrs}$.
We identify unsafe $k$ by slicing $Z\frs\idxfrs$ and checking if it intersects $O$.
Suppose $G\frs\idxfrs$ has $n \in \N$ generators.
Let $I = \N_{\Kdim} + \plandim$.
Denote
\begin{align*}
    G\slc\idxfrs &= [G\frs\idxfrs]\arridx{\N_{\plandim},I},~G\buf\idxfrs = [G\frs\idxfrs]\arridx{\N_{\plandim},\N_n\setminus I},\\
    Z\slc\idxfrs &= \zono{c\frs\idxfrs,G\slc\idxfrs},~Z\buf\idxfrs = \zono{0,G\buf\idxfrs}.
\end{align*}
Let $k = (k\ini,k\desi) \in K\idxj$ such that $k\ini = f\ini(x\hiz)$, and construct $\beta_k = \diag{\Kdelta}\inv(k - \Kcenter)$.
Then
\begin{align}
    \slicefunc\left(Z\frs\idxfrs,\beta_k,I\right)~\cap~O = \emptyset 
\end{align}
if and only if
\begin{align}
    \slicefunc\left(Z\slc\idxfrs,\beta_k,\N_{\Kdim}\right)~\not\in~\zono{c\obs\idxm,[G\obs\idxm,G\buf\idxfrs]}.\label{eq:eval_slice_zono_Z_slc}
\end{align}
\end{lem}
\begin{proof}
First note, $O$ is a rotated box, and therefore a zonotope.
Second, notice that, by construction, $Z\slc\idxfrs$ and $Z\buf\idxfrs \subset P$.
Furthermore, $Z\slc\idxfrs$ contains the generators of $Z\frs\idxfrs$ that can be sliced by $\beta_k$ \cite[Lemma 6.5]{shreyas_dissertation}, and $Z\buf\idxfrs$ contains all of the other generators (hence the multi-index $\N_n \setminus I$).
This means the left-hand side of \eqref{eq:eval_slice_zono_Z_slc} is a point, hence the use of $\not\in$.
Furthermore, it follows from \eqref{eq:slice_defn} that $\slicefunc\left(Z\slc\idxfrs,\beta_k,\N_{\Kdim}\right) \in P$, because $Z\slc\idxfrs$ has exactly $\N_\Kdim$ generators \cite[Lemma 6.5]{shreyas_dissertation}.
The desired result then follows from \cite[Lemma 5.1]{guibas2003zonotopes}: if $Z_1 = \zono{c_1,G_1}$ and $Z_2 = \zono{c_2, G_2}$, then $Z_1 \cap Z_2 = \emptyset \iff c_1 \not\in \zono{c_2,[G_1,G_2]}$.
\end{proof}


In our application, we must repeatedly evaluate \eqref{eq:eval_slice_zono_Z_slc} (see Alg. \ref{algo:adjust}, Line \ref{algo:adjust:line:check_safety_loop}).
To perform this efficiently, we apply \cite[Thm. 2.1]{althoff2010reachability} to represent each zonotope $\zono{c\obs\idxm,[G\obs\idxm,G\buf\idxfrs]}$ as the intersection of $n_{\regtext{hp}} \in \N$ affine halfplanes using a pair of matrices, $A\obs\idxhp \in \R^{n_{\regtext{hp}}\times \plandim}$ and $\quad b\obs\idxhp \in \R^{n_{\regtext{hp}}\times 1}$, for which \eqref{eq:eval_slice_zono_Z_slc} holds if and only if
\begin{align}\label{eq:halfplane_safety_check}
  -\max\left(A\obs\idxhp\,\slicefunc(Z\slc,\beta_k,\N_{\Kdim}) - b\obs\idxhp\right) < 0,  
\end{align}
where the max is taken over the elements of its argument.
Note $A\obs\idxhp$ and $b\obs\idxhp$ can be constructed quickly, and enable future work where the $\adjustfunc$ function in Alg. \ref{algo:adjust} can use gradient descent instead of sampling, similar to \cite{holmes2020reachable}.

\subsubsection{Parameter Adjustment}
To enforce safety at runtime, we use \eqref{eq:halfplane_safety_check} as a constraint on the RL agent's choice of $k$.
Using Alg. \ref{algo:adjust}, we adjust an unsafe choice of $k$ by attempting to replace it with a safe one.
Importantly, Alg. \ref{algo:adjust} also returns the Euclidean distance from the RL agent's choice to the adjusted $k$, which we can use as a penalty during training.

\begin{algorithm}\label{algo:adjust}
\caption{$(k,d) = \adjustfunc(x\hiz,\{O\idxm\}_{m=1}^{n\obs},k\RL)$}
\SetAlgoVlined
get $Z\plan\idxplan$ and $Z\err\idxerr$ for each $i \in \N_{\numT}$ using $x\hiz$

create $Z\frs\idxfrs$ for each $i \in \N_{\numT}$ as in \eqref{eq:ERS_zono_plus_PRS_zono}

\For{$i \in \N_{\numT}, m \in \N_{n\obs}$}{
    construct $A\obs\idxhp$ and $b\obs\idxhp$ as in Sec. \ref{subsec:creating_safety_cons}\\
    
    evaluate \eqref{eq:halfplane_safety_check} on $k\RL$
}

\eIf{$k\RL$ does not satisfy \eqref{eq:halfplane_safety_check} for all $(i,m)$}{
    create samples $\{k\idxn\}_{n=1}^{n\sample} \subset K\idxj$ such that $k\ini\idxn = f\ini(x\hiz)$ for all $m$
    
    sort $\{k\idxn\}_{n=1}^{n\sample}$ by increasing cost $\Vert k\idxn - k\RL\Vert_2$
    
    \For{$i \in \N_{\numT}$, $m \in \N_{n\obs}$, $n \in \N_{n\sample}$ \label{algo:adjust:line:check_safety_loop}}{
        \If{$k\idxn$ satisfies \eqref{eq:halfplane_safety_check}}{
        \textbf{return} $(k\idxn ,\Vert k\idxn - k\RL\Vert_2)$, \textbf{break} loop
        }
    }
    
    \textbf{return} $([\ ],[\ ])$ (robot continues previous plan)
    
}{
    \textbf{return} $(k\RL,0)$ (no adjustment necessary)
}
\end{algorithm}
\subsection{Safe Learning with RTS}
We use RTS to safely train a model-free RL agent with Alg. \ref{algo:saferl}.
In each training episode the RL agent performs receding-horizon planning until the robot completes the task (e.g., reaching a goal), crashes, or exceeds a time limit.
In each planning iteration, we roll out the current policy to get a plan, adjust the plan if unsafe, and execute the resulting plan.
We train the RL agent on minibatches of \defemph{experiences} containing observations, reward, and policy output.
The \defemph{observations} contain the robot's state, nearby obstacles, and goal information.
The \defemph{reward} is a function of the task, robot trajectory, obstacles, and distance that Alg. \ref{algo:adjust} adjusted the agent's plan.
In practice, training the agent at runtime does not impede the real-time performance of RTS because we use an experience buffer, allowing training in parallel to plan execution.
We conclude by confirming that RTS is safe:

\begin{thm}\label{thm:safe_RL}
Suppose the ERS satisfies \eqref{eq:ers_conservatism}.
Then an RL agent / robot using Alg. \ref{algo:saferl} is safe during training.
\end{thm}
\begin{proof}
In each planning iteration, Alg. \ref{algo:saferl} checks if the plan from the RL agent is unsafe using Alg. \ref{algo:adjust}, which either returns a safe plan (by Lem. \ref{lem:test_if_sliced_FRS_intersects_obs}), or else the robot executes a previously-found, safe failsafe maneuver.
Since the robot is initialized with a failsafe maneuver, it is always safe.
\end{proof}


\begin{algorithm}
\caption{Safe Reinforcement Learning with RTS}\label{algo:saferl}
\SetAlgoVlined
initialize random policy and empty experience set $\mathcal{E}$
  
\For{each training episode}{
    initialize task $\mathcal{T}$ (e.g., reach a goal position)
    
    create random obstacles $\obsset = \{O\idxm\}_{m=1}^{n\obs}$ \label{algo:saferl:line:episode_setup}
    
    initialize robot at random, safe start position, with failsafe maneuver (stay at start)
    
    \While{time limit not exceeded}{
        get initial condition $x\hiz$ from robot
        
        get observation $o = \observationfunc(x\hiz,\obsset)$\label{algo:saferl:line:obs_before}
    
        get $k\RL = \rolloutfunc(o)$ \label{algo:saferl:line:rollout}
        
        get $(k\safe,d) = \adjustfunc(x\hiz,\obsset,k\RL)$ with Alg. \ref{algo:adjust}\label{algo:saferl:line:adjust}
        
        \eIf{$k\safe \neq [\ ]$}{
            execute robot trajectory $x\hi$ as in \eqref{eq:hi_fid_model_traj_with_u_trk} by tracking $p\plan(\cdot,k\safe)$ for $t\plan$ s
            
            store $p\plan$ as new failsafe maneuver
        }{
            execute robot trajectory $x\hi$ as in \eqref{eq:hi_fid_model_traj_with_u_trk} by continuing previous failsafe for $t\plan$ s
        }
        
        get $r = \rewardfunc(\mathcal{T},x\hi,\obsset,d)$\label{algo:saferl:line:get_reward}
        
        get observation $o' = \observationfunc(x\hi(t\plan),\obsset)$\label{algo:saferl:line:obs_after}
        
        store experience $(o,o',r,k\RL)$ in $\mathcal{E}$\label{algo:saferl:line:store_exp}
        
        train policy on a minibatch of $\mathcal{E}$ \label{algo:saferl:line:train}
        
        \textbf{if} \textit{task $\mathcal{T}$ done or robot crashed} \textbf{then} break\label{algo:saferl:line:check_if_done}
    }
}
\Return trained policy
\end{algorithm}


\section{Experiments}\label{sec:experiments}


The proposed approach is demonstrated on a cartpole robot, an autonomous car, and a quadrotor drone, all simulated in MATLAB.
Due to space limitations, results for the cartpole and implementation details for all robots are in the supplement.
A \href{https://youtu.be/j5h3JzHboMk}{\textcolor{blue}{\underline{supplementary video}}} highlights our method.

\subsubsection*{Comparison Methods}
For each robot, we train three RL agents: one with RTS to ensure safety, one with RTS but a discrete action space (similar to \cite{TUM_safe_lane_change_set_based}), and a baseline with no safety layer.
We also compare against two versions of RTD \cite{shreyas_dissertation}: a ``Reward'' version that optimizes the same reward as RL, and a ``Standard'' version that optimizes distance to a waypoint generated by a high-level path planner.
At the time of writing, code to compare against \cite{fisac2018general} was unavailable.

\subsubsection*{Evaluation Metrics}
We consider goals reached (i.e., tasks completed), safe stops (task incomplete, but no collision), collisions, safety interventions (how many times Alg. \ref{algo:adjust} was needed), and min/mean/max reward over all trials.
To ensure a fair comparison, all methods run for a fixed number of planning iterations; when a method gets stuck, it accumulates negative reward per the reward functions in the supplement.

\subsubsection*{Hypotheses}
We expect RTS+RL and RTD to have no collisions, the continuous action space RTS to outperform the discrete action space version, and baseline RL to have many collisions.
We expect Standard RTD to outperform Reward RTD due its convex cost and hand-tuned high-level planner.

\subsubsection*{Summary of Results}
RTS+RL consistently outperforms the other methods on most metrics.
Standard RTD consistently outperforms Reward RTD as expected.
Interestingly, Standard RTD often achieves higher reward (but fewer goals) than RTS for the drone, meaning reward is not always an accurate metric for task success.

\subsection{Car Lane Change Experiment}
\subsubsection{Task and Method}
In this experiment, a self-driving car tries to reach a goal position 500 m away on a road-like obstacle course as quickly as possible.
We use a realistic high-fidelity model \cite{rasekhipour2016potential} that has a larger turning radius at higher speeds, so the car must slow to avoid obstacles, and stop if there is not enough room to avoid an obstacle.
For the RL methods, we train TD3 \cite{fujimoto2018addressing} agents for 20,000 episodes and evaluate on 500 episodes.


\subsubsection{Results and Discussion}
\begin{figure}[t]
  \centering
  \includegraphics[width=0.8\linewidth]{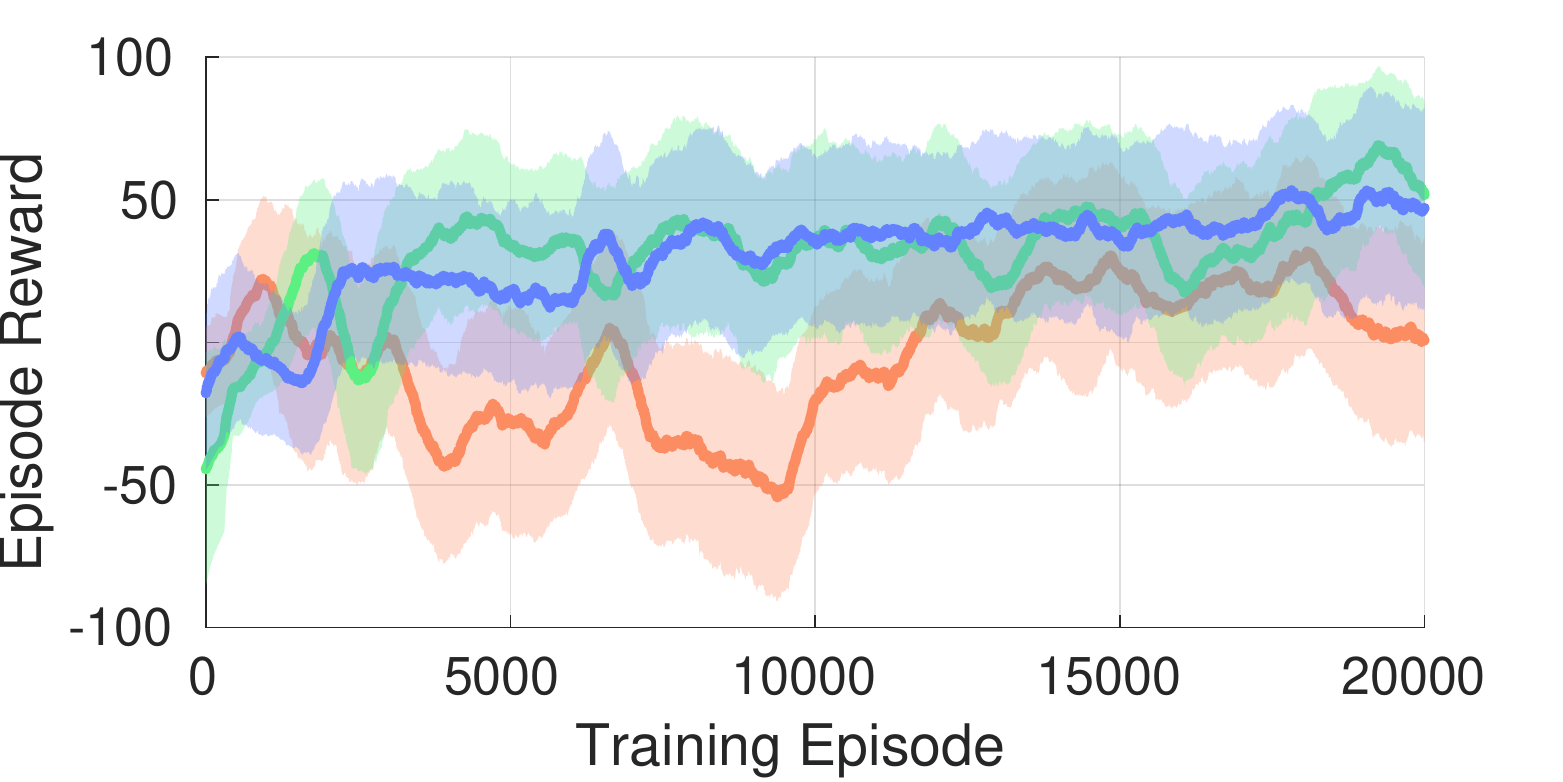}
  \caption{Running average of reward and its standard deviation during training for the car lane change task.
  The proposed RTS+RL method (green) achieves high reward compared to a discrete version of the same method (blue) and a baseline RL approach (orange).
  Baseline RL had collisions in 40.2\% of episodes, whereas RTS had none.}
  \label{fig:train_car}
\end{figure}

\begin{figure}[t]
    \centering
    \includegraphics[width=0.95\linewidth]{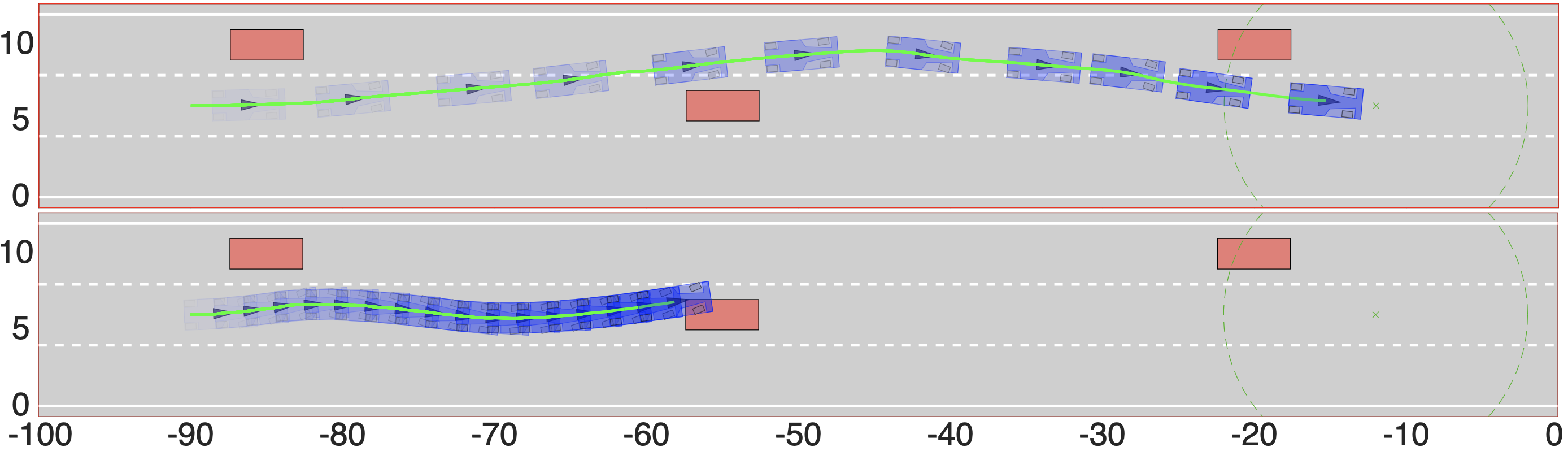}
    \label{fig:car_sim_N}
    \caption{Car lane changes with RTS+RL (top) and baseline RL (bottom).
    The car (blue) is plotted at each receding-horizon planning iteration (increasing opacity with time).
    RTS+RL avoids obstacles (red) while traveling at a higher speed than the baseline RL agent, which suffers a collision.}
    \label{fig:car_sim}
    \vspace*{-0.2cm}
\end{figure}

Fig. \ref{fig:train_car} shows reward during training.
Table \ref{tbl:car} shows evaluation data.
Fig. \ref{fig:car_sim} illustrates RTS+RL achieving two safe lane changes at high speed, and the baseline RL agent having a collision.

RTS+RL attains high reward and goals by learning to drive slowly near obstacles and quickly otherwise.
RTS+RL Discrete is less consistent because it lacks fine control over the car's speed, limiting possible turning radii and getting the car stuck.
Baseline RL collides often, as expected, so it learns to drive slowly, limiting its reward.
As expected, RTD does not crash, and Standard RTD outperforms Reward RTD.
Standard RTD succeeds due to our careful hand-tuning of its high-level planner, resulting in similar success rates to prior studies on RTD \cite{vaskov2019towards,kousik2018bridging}.
However, RTS effectively behaves as an automated way of tuning this high-level behavior, resulting in a higher success rate with less human effort (we found tuning the reward function easy in practice since there is no tradeoff required for penalizing obstacles/collisions).

Note, RTS requires much less than $t\plan = 2$ seconds to ensure safety in each planning iteration, meaning that it enables real-time training and evaluation.
We chose this $t\plan$ because it forces the methods to consider longer-term reward and discourages aggressive lateral acceleration
Also, the dynamics we use for RTS+RL have been shown to accurately represent real car-like robots for safe operation \cite{kousik2018bridging,vaskov2019towards}, so we expect that RTS can overcome the sim-to-real gap.

\begin{table*}[t]
\centering
\begin{tabular}{l|r|r|r|r|r}
\rowcolor[HTML]{C0C0C0} 
{\textbf{Car Results}} & \multicolumn{1}{c|}{\cellcolor[HTML]{C0C0C0}{\textbf{RTS+RL}}} & \multicolumn{1}{c|}{\cellcolor[HTML]{C0C0C0}{\textbf{RTS+RL Discrete}}} & \multicolumn{1}{c|}{\cellcolor[HTML]{C0C0C0}\textbf{Baseline RL}} & \multicolumn{1}{c|}{\cellcolor[HTML]{C0C0C0}\textbf{Reward RTD}} & \multicolumn{1}{c}{\cellcolor[HTML]{C0C0C0}\textbf{Standard RTD}} \\ \hline
\rowcolor[HTML]{FFFFFF} 
Avg. Planning Time {[}s{]}                              & 0.058                                                                               & 0.057                                                                                        & \textbf{1.7E-5}                                                   & 0.20                                                            & 0.17                                                            \\ \hline
\rowcolor[HTML]{FFFFFF} 
Goals Reached {[}\%{]}                                  & \textbf{100.0}                                                                      & 88.0                                                                                         & 82.4                                                              & 42.8                                                            & 90.2                                                            \\
\rowcolor[HTML]{FFFFFF} 
Safely Stopped {[}\%{]}                                 & \textbf{0.0}                                                                       & 12.0                                                                                         & \textbf{0.0}                                                      & 57.2                                                            & 9.8                                                            \\
\rowcolor[HTML]{FFFFFF} 
Collisions {[}\%{]}                                     & \textbf{0.0}                                                                        & \textbf{0.0}                                                                                 & 17.6                                                              & \textbf{0.0}                                                    & \textbf{0.0}                                                    \\

\rowcolor[HTML]{FFFFFF} 
Safety Interventions {[}\%{]}                           & \textbf{3.3}                                                                      & 6.2                                                                                        & N/A                                                               & N/A                                                             & N/A  \\  \hline
\rowcolor[HTML]{FFFFFF} 
Min/Mean/Max Reward & $\textbf{38} / \mathbf{121} / \textbf{158}$ & $-645 / 87 / 157$ & $-53 / 78 / 151$ & $-1471 / -168 / 156$ & $-821 / 53 / 157$ 

\end{tabular}
\caption{Evaluation and comparison results for the Car Lane Change Experiment.
Reward is rounded to nearest integer for space.
}
\label{tbl:car}
\end{table*}

\subsection{Drone Obstacle Tunnel Experiment}

\subsubsection{Task and Method}
\begin{figure}[t]
    \centering
    \includegraphics[trim={7cm 0cm 7cm 0cm},clip,width=\linewidth,]{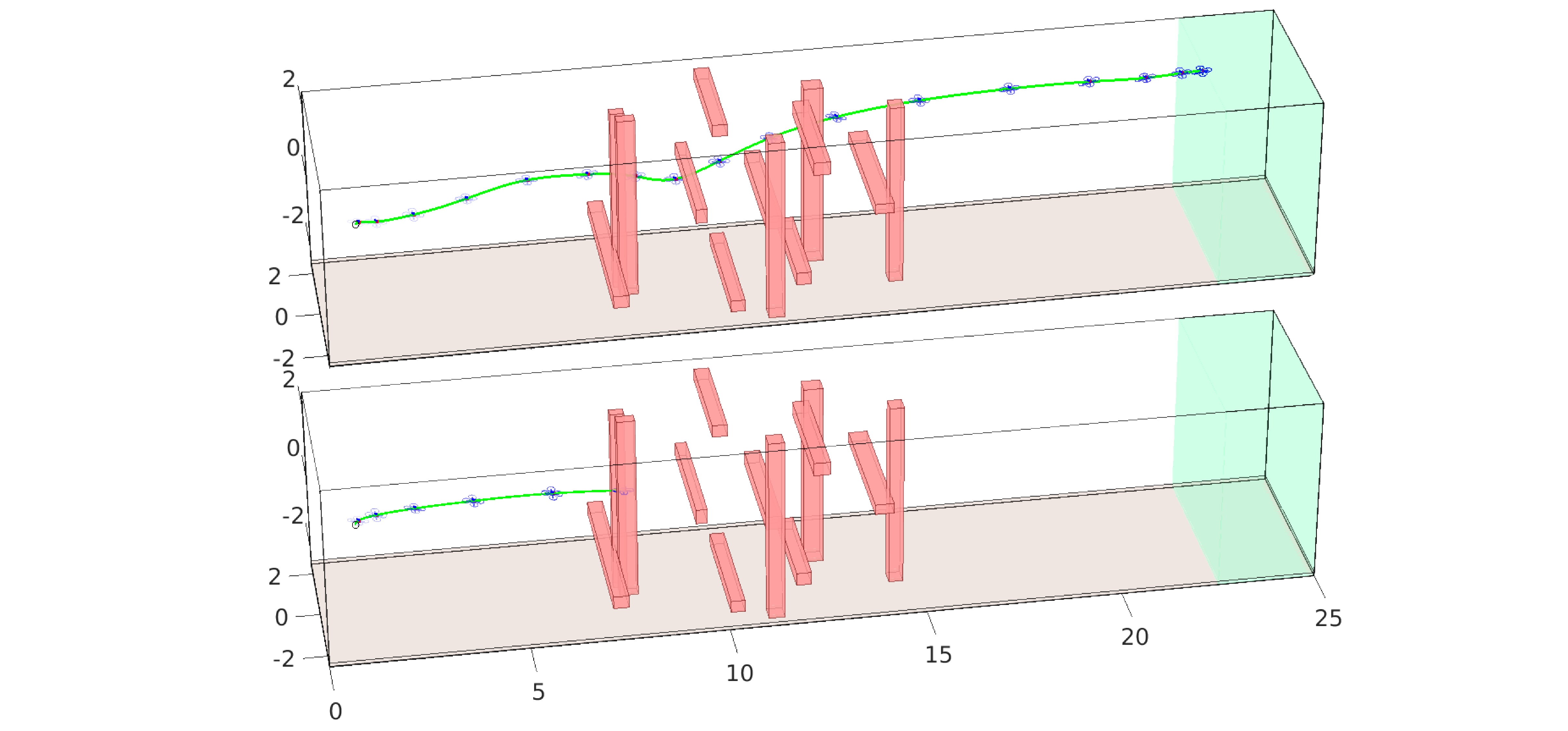}
    \caption{In the Drone Obstacle Tunnel Experiment, RTS+RL (top) successfully navigates a trial, whereas the baseline RL (bottom) learns to crash rapidly to avoid accumulating negative reward.
    The drone begins on the left and must reach the goal (green) on the right while avoiding obstacles (red).
    Note, unlike this 25 m example world, the training and testing worlds are 100 m long and have higher obstacle density.}
    \label{fig:drone_sim}
    \vspace*{-0.3cm}
\end{figure}

This experiment requires a quadrotor drone to traverse a 100 m tunnel as quickly as possible while avoiding randomly-placed obstacles, as shown in Fig. \ref{fig:drone_sim}.
Recent applications of deep/reinforcement learning for drone control and navigation have only empirically demonstrated safety of a learned policy \cite{hwangbo2017control,waslander2005multi,kaufmann2020deep}.
We use RL+RTS to enable more systematic guarantees for learning drone navigation.
We train TD3 agents for 2000 episodes, then evaluate on 500 episodes.


\begin{figure}[t]
  \centering
  \includegraphics[width=0.8\linewidth]{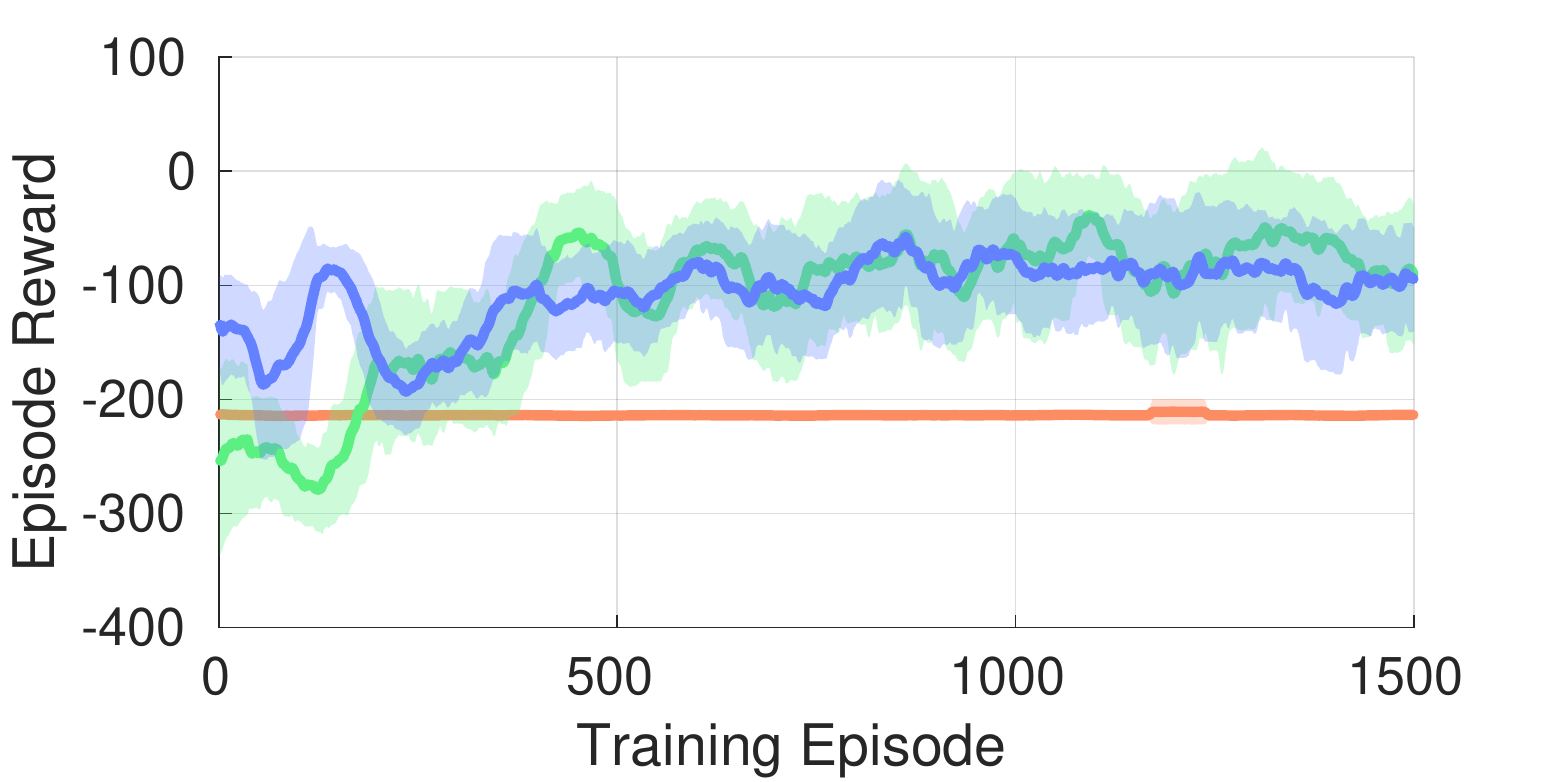}  \caption{Running average of reward and its standard deviation during training for the drone obstacle tunnel.
  Our RTS+RL framework (green) learns to navigate the random obstacle tunnels, whereas the discrete version (blue) does not achieve as much reward.
  The baseline RL approach (orange) struggles to accumulate reward, and instead learns to collide with obstacles}
  \label{fig:train_drone}
  \vspace*{-0.2cm}
\end{figure}

\subsubsection{Results and Discussion}
Table \ref{tbl:drone} shows evaluation data.
Fig. \ref{fig:drone_sim} shows an example where RTS+RL succeeds and baseline RL has a collision.
As expected, RTS+RL and RTD had no collisions, and the continuous action space RTS was superior.
The discrete action space often has too few actions to prevent the robot from becoming stuck.
Baseline RL learned to crash to avoid accumulating negative reward over time; so, enforcing safety allows one to avoid some tradeoffs in reward tuning.
As expected, Standard RTD was more effective than Reward RTD at reaching goals and accumulating reward due to its carefully hand-tuned high level planner.
Surprisingly, RTS+RL reached more goals (which is RTD's purpose) but RTD accumulated higher reward, showing that reward is not necessarily the proper metric for evaluating an RL agent.
Also, note RTS+RL's planning time is less than the real-time limit $t\plan = 1$ s.

\begin{table*}[t]
\centering
\begin{tabular}{l|r|r|r|r|r}
\rowcolor[HTML]{C0C0C0} 
\multicolumn{1}{l|}{\cellcolor[HTML]{C0C0C0}\textbf{Drone Results}} & \multicolumn{1}{c|}{\cellcolor[HTML]{C0C0C0}\textbf{RTS+RL}} & \multicolumn{1}{c|}{\cellcolor[HTML]{C0C0C0}\textbf{RTS+RL Discrete}} & \multicolumn{1}{c|}{\cellcolor[HTML]{C0C0C0}\textbf{Baseline RL}} & \multicolumn{1}{c|}{\cellcolor[HTML]{C0C0C0}\textbf{Reward RTD}} & \multicolumn{1}{c}{\cellcolor[HTML]{C0C0C0}\textbf{Standard RTD}} \\ \hline
\rowcolor[HTML]{FFFFFF} 
Avg. Planning Time {[}s{]}                                                          & 0.22                                                         & 0.18                                                                  & \textbf{8.8E-6}                                                   & 0.86                                                                & 0.22                                                              \\ \hline
\rowcolor[HTML]{FFFFFF} 
Goals Reached {[}\%{]}                                                              & \textbf{83.4}                                                & 71.8                                                                  & 0.0                                                               & 58.6                                                                & 76.2                                                              \\
\rowcolor[HTML]{FFFFFF} 
Safely Stopped {[}\%{]}                                                             & \textbf{16.6}                                                & 28.2                                                                  & 0.0                                                               & 41.4                                                                & 23.8                                                              \\
\rowcolor[HTML]{FFFFFF} 
Collisions {[}\%{]}                                                                 & \textbf{0.0}                                                 & \textbf{0.0}                                                          & 100.0                                                             & \textbf{0.0}                                                        & \textbf{0.0}                                                      \\
\rowcolor[HTML]{FFFFFF} 
Safety Interventions {[}\%{]}                                                       & 90.6                                                         & \textbf{80.3}                                                         & N/A                                                               & N/A                                                                 & N/A                                                               \\ \hline
\rowcolor[HTML]{FFFFFF} 
Min/Mean/Max Reward  & $-430/-63/25$ & $-429/-95/30$ & $\mathbf{-212}/{-212}/{-210}$ & ${-345}/-112/31$ & $-430/\mathbf{-55}/\textbf{43}$ 
\end{tabular}
\caption{Evaluation and comparison results for the Drone Obstacle Tunnel Experiment.
Reward is rounded to nearest integer for space.}
\label{tbl:drone}
\vspace*{-0.5cm}
\end{table*}

\section{Conclusion}\label{sec:conclusion}
To apply RL on real-world robots in safety-critical environments, one should be able to ensure safety during and after training.
To that end, this paper proposes Reachability-based Trajectory Safeguard (RTS), which leverages offline reachability analysis to guarantee safety.
The method is demonstrated in simulation performing safe, real-time receding-horizon planning for three robot platforms with continuous action spaces.
RTS typically outperforms state-of-the-art safe trajectory planners in terms of reward and tasks completed.
Furthermore, RTS simplifies RL training by allowing users to focus on designing a reward without tuning safety penalties.
Future work will apply RTS+RL on hardware and non-rigid-body robots, and explore additional benefits of safe RL training.



\renewcommand{\bibfont}{\normalfont\footnotesize}
{\renewcommand{\markboth}[2]{}
\printbibliography}

\clearpage
\setcounter{page}{1}
\setcounter{section}{0}
\section*{RTS Supplement}\label{app}
In this supplement, we provide the following\footnote{Supplementary video: \url{https://youtu.be/j5h3JzHboMk}}.
First, we elaborate on the utility of slicing the FRS.
Second, we provide the cartpole experiment results, and implementation details for the cartpole, car, and quadrotor drone.
In particular, we present the high-fidelity model, planning model, tracking controller, and RL reward function for each robot.

\section{Slicing the FRS Zonotopes}
We now explain slicing in more detail.
For context, slicing is a key difference between our use of reachable sets and that of, e.g., \cite{TUM_safe_lane_change_set_based,hess2014_parameterized_motion_primitives}.
In particular, slicing allows us to back out a subset of a high-dimensional reachable set by only considering a single trajectory parameter, as opposed to the entire space of trajectory parameters.
To enable slicing, our reachable sets include the parameters as augmented state dimensions, in contrast to \cite{hess2014_parameterized_motion_primitives} where the reachable sets are effectively unions (in state space) over every parameterized trajectory.
Our reachable sets are instead \emph{disjoint} unions (in state space and parameter space) of all of reachable sets of each parameterized trajectory.

We find that slicing, in combination with our sampling-based reachability analysis, significantly reduces the conservatism and computation time of using overapproximative reachable sets for online planning.
This approach also enables extensions to high-dimensional nonlinear robot models such as the quadrotor in the present work.

To demonstrate the utility of slicing, we check that the FRS does indeed contain the motion of the robot when tracking any plan:
\begin{lem}\label{lem:sliced_FRS_is_conservative}
(\cite[Thm. 6.6]{shreyas_dissertation})
Suppose $t \in T\idxi$.
Let $k \in K\idxj$.
Suppose $Z\err\idxerr$ satisfies \eqref{eq:ers_conservatism}.
Let
\begin{align}\label{eq:beta_k_slice_coeff_defn}
    \beta_k = \diag{\Kdelta}\inv(k - \Kcenter).
\end{align}
Let $I = \N_\Kdim + \plandim$.
Then
\begin{align}
    \forwardoccupancy\left(x\hi(t;x\hiz,k)\right) \subset \slicefunc(Z\frs\idxfrs,\beta_k,\N_\Kdim).
\end{align}
\end{lem}
\noindent Note, $I$ is constructed in this way because the first $\plandim$ columns of the generator matrix correspond to the $p\plan$ dimensions of $Z\frs\idxfrs$; the next $\Kdim$ columns correspond to $k$ \cite[Lemma 6.5]{shreyas_dissertation}, and all other columns correspond to the ERS zonotope by construction.
So, we slice all the generators that correspond to $k$; the remaining generators add volume to $Z\frs\idxfrs$ to compensate for tracking error, the body of the robot, and nonlinearities in $\dot{p}\plan(\cdot,k)$.

In short, one can slice the FRS to find the reachable volume in workspace corresponding to a particular plan, as illustrated in Fig. \ref{fig:PRS} (note, this figure just shows the sliced PRS for visual clarity).
The plan is safe if this volume does not intersect with obstacles.

\section{Experiments}

We now present the cartpole experiment, with a comparison of RTS+RL against baseline RL and RTD with the RL reward function. 
We then provide implementation details for the car and quadrotor drone\footnote{Our code is available online at \url{www.github.com/roahmlab/reachability-based_trajectory_safeguard}}.

\subsection{Cartpole Implementation Details}

To demonstrate safe RL on a simple example, we use a cartpole, or an unactuated pendulum on a cart.
We consider the swingup task, wherein one must invert the pendulum by moving the cart.
We limit the length of track, and seek to complete the task without exceeding the track boundaries.
We do not consider the pendulum for obstacle avoidance (to keep the example simple), but the method can extend to include the pendulum using rotating body reachable sets as in \cite{holmes2020reachable}.

\subsubsection{High-Fidelity Model}
The cartpole's state is $x\hi = (p,\dot{p},\theta,\dot{\theta})$, containing the cart position and velocity ($p$ and $\dot{p}$) and the pendulum angle (relative to vertical) and velocity ($\theta$ and $\dot{\theta}$).
We use the following high-fidelity model \cite{Pati2014ModelingIA}:
\begin{align}
\ddot{p} &=\frac{\left(w + m l^{2}\right)\left(u+m l \dot{\theta}^{2} \sin \theta\right)-g m^{2} l^{2} \sin \theta \cos \theta}{w(m\cart+m)+m l^{2}\left(m\cart+m \sin ^{2} \theta\right)} \\
\ddot{\theta} &= \frac{-m l(u \cos \theta+m l \dot{\theta}^{2} \sin \theta \cdot \cos \theta-(m\cart+m) g \sin \theta)}{w(m\cart+m)+m l^{2}\left(m\cart+m \sin ^{2} \theta\right)},
\end{align}
where $w = 0.099$ kg$\cdot$m\ts{2} is the inertia of the pendulum, $m = 0.2$ kg is the mass of pole, $m\cart = 2$ kg is the mass of the cart, and $l = 0.5$ m is the length of the pole \cite[Table I, pg. 8]{Pati2014ModelingIA}.
The control input is $u$.

For $X\hi$, we require $p(t) \in [-4,4]$ m (instead of randomizing the track length, we randomize the cart and pendulum initial state in each episode).
We require $\dot{p}(t) \in [-5,5]$ m/s, $\theta \in [-\pi,\pi]$ rad.
We do not explicitly bound $\dot{\theta}$.
We draw control inputs from $U = [-40,40]$ N.

\subsubsection{Planning Model} \label{app:cartpole_planning}
We use a planning model based on the 1-D model in \cite{mueller_drone_motion,kousik2019safe}.
The trajectory parameters are $(k_v,k_a) \in K\ini \subset \R^2$ and $k\peak \in K\desi \subset \R$; $k_v$ (resp. $k_a$) is the cart's initial velocity (resp. initial acceleration), and $k\peak$ is a desired velocity to be achieved at a time $t\peak \in (0,t\final)$.
The planning model is:
\begin{align}
p\plan(t, k) &= \frac{\tau_1(t, k)}{24} t^{4} + \frac{\tau_2(t, k)}{6} t^{3} + \frac{k_{a}}{2} t^2 + k_{v}t \label{eq:ref_cartpole}\\
\left[\begin{array}{c}
\tau_1(t, k) \\
\tau_2(t, k)
\end{array}\right] &= \frac{1}{\left(\tau_3(t)\right)^{3}}\left[\begin{array}{cc}
-12 & 6 \tau_3(t) \\
6 \tau_3(t) & -2\left(\tau_3(t)\right)^{2}
\end{array}\right]\left[\begin{array}{c}
\Delta_{v}(t, k) \\
\Delta_{a}(t, k)
\end{array}\right] \\
\tau_3(t) &= \left\{\begin{array}{ll}
t\peak & t \in\left[0, t\peak\right) \\
t\final-t\peak & t \in\left[t\peak, t\final\right]
\end{array}\right.\\
\Delta_{v}(t, k) &= \left\{\begin{array}{ll}
k\peak-k_{v}-k_{a} t\peak & t \in\left[0, t\peak\right) \\
-k\peak & t \in\left[t\peak, t\final\right]
\end{array}\right.\\
\Delta_{a}(t, k) &= \left\{\begin{array}{ll}
-k_{a} & t \in\left[0, t\peak\right) \\
0 & t \in\left[t\peak, t\final\right]
\end{array}\right.
\end{align}
We set $k_v, k\peak \in [-5,5]$ m/s and $k_a \in [-15,15]$ m/s\ts{2}.
We set $t\plan = t\peak = 0.1$ s and $t\final = 0.3$ s.

\subsubsection{Tracking  Controller} \label{app:cartpole_tracking}
We use
\begin{align}
u\track(t,x\hi(t),k) = \big[\gmp, \gmd\big]\cdot\begin{bmatrix}
    p\plan(t,k) - p(t) \\
    \dot{p}\plan(t,k) - \dot{p}(t)
\end{bmatrix},
\end{align}
where $\gmd = \gmp = 50$ are control gains with the appropriate units.
We saturate this controller if $|u\track(\cdot)| > 40$ N.

\subsubsection{Reachability Hyperparameters}
To compute the PRS and ERS, we partition $T$ into $\numT = 30$ intervals.
We partition the $k_v$ dimension of $K$ into $11$ intervals and $k_a$ into $5$ intervals, so $\numK = 11\times5$.
We similarly partition the $\dot{p}$ dimension of $X\hiz$ into 11 intervals; we partition the $\theta$ dimension into 4 intervals, and do not partition the $\dot{\theta}$ dimension, so $\numXhi = 11\times 4$.
We found that, since the pendulum is light compared to the cart, the cart's tracking error is not significantly influenced by the pendulum's speed.
Also note, we do not need to partition the $p$ dimension of $X\hiz$ because all plans start at $p(0) = 0$.

\subsubsection{Observations}
We provide the RL agent with observations of the robot's state: $o = (p,\dot{p},\sin\theta,\cos\theta,\dot{\theta})$.

\subsubsection{Reward} \label{app:cartpole_reward}
We specify three reward terms: $r_1$ rewards bringing the pendulum upright, $r_2$ rewards being in the middle of the track, and $r_3$ penalizes exceeding track boundaries.
Recall that $x\hi = (p,\dot{p},\theta,\dot{\theta}) \in X\hi$.
\begin{align}
    \regtext{r}(x\hi) &= r_1(x\hi) + r_2(x\hi) + r_3(x\hi),~\regtext{where}\\
   \regtext{r}_1(x\hi) &= \tfrac{1}{2}\cos\theta + \tfrac{1}{2} \\
   \regtext{r}_2(x\hi) &= - 0.1\cdot\sign(p)\cdot\sign(\dot p),\ \regtext{and}\\
   \regtext{r}_3(x\hi) &= \begin{cases}
    -0.05 |p| + 30 & p \in [-4,4]\\
    -0.05 |p| - 30 & p \notin [-4,4].
   \end{cases}
\end{align}
Here, $\sign$ returns $+1$ or $-1$ (i.e., the sign of its argument).

\subsection{Cartpole Swing Up Experiment}

\begin{figure}[t]
    \raggedleft
    \includegraphics[trim={2.7cm 0.6cm 2.2cm 1 cm},clip,width=1.02\linewidth]{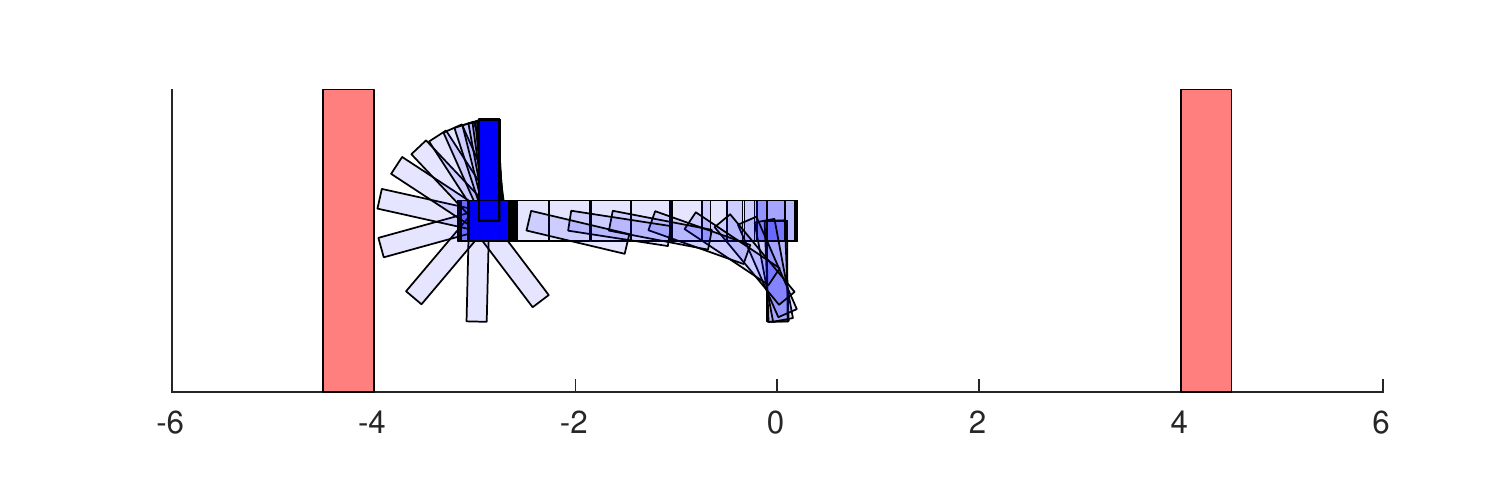}
    \caption{A time lapse illustration of the cartpole swingup policy learned by our proposed RTS+RL method.
    The cart is the blue rectangle that moves horizontally, and the pendulum is the blue rectangle that rotates; the red walls are the track boundaries; the track is not shown to reduce visual clutter.
    The cart begins at $0$ and moves to the left, then brakes to a stop before hitting the boundary while still completing the swingup motion.
    This figure shows a hand-crafted initial condition to emphasize our approach's ability to obey the constraints; the training and evaluation episodes had randomized initial conditions, but RTS+RL still obeyed the constraints.}
    \label{fig:cartpole_sim}
\end{figure}

\subsubsection{Task, Method, and Hypotheses}
The cartpole task is to swing up the freely-rotating, unactuated pendulum by planning trajectories for the cart while ensuring safety by staying within the track bounds.

We train two DDPG \cite{lillicrap2015continuous} RL agents (one with RTS and one without, as noted in Sec. \ref{sec:experiments}) for 300 episodes, and is tested 500 episodes.
Every episode starts at a random initial state, with the cart position far enough from the track limits that it is possible to avoid collision.
We also test Reward RTD (i.e. optimizing the RL reward).
A Standard RTD approach, with a high-level planner to generate waypoints towards the global goal, does not exist for the cartpole.

Since RL optimizes for the long-term reward, we expect both RL agents to attempt to complete the task.
We expect RTS will always respect the track limits, unlike baseline RL.
Finally, we expect RTD to complete the task without a high-level planner.

\subsubsection{Results and Discussion}
\begin{figure}[t]
  \centering
  \includegraphics[width=0.9\linewidth]{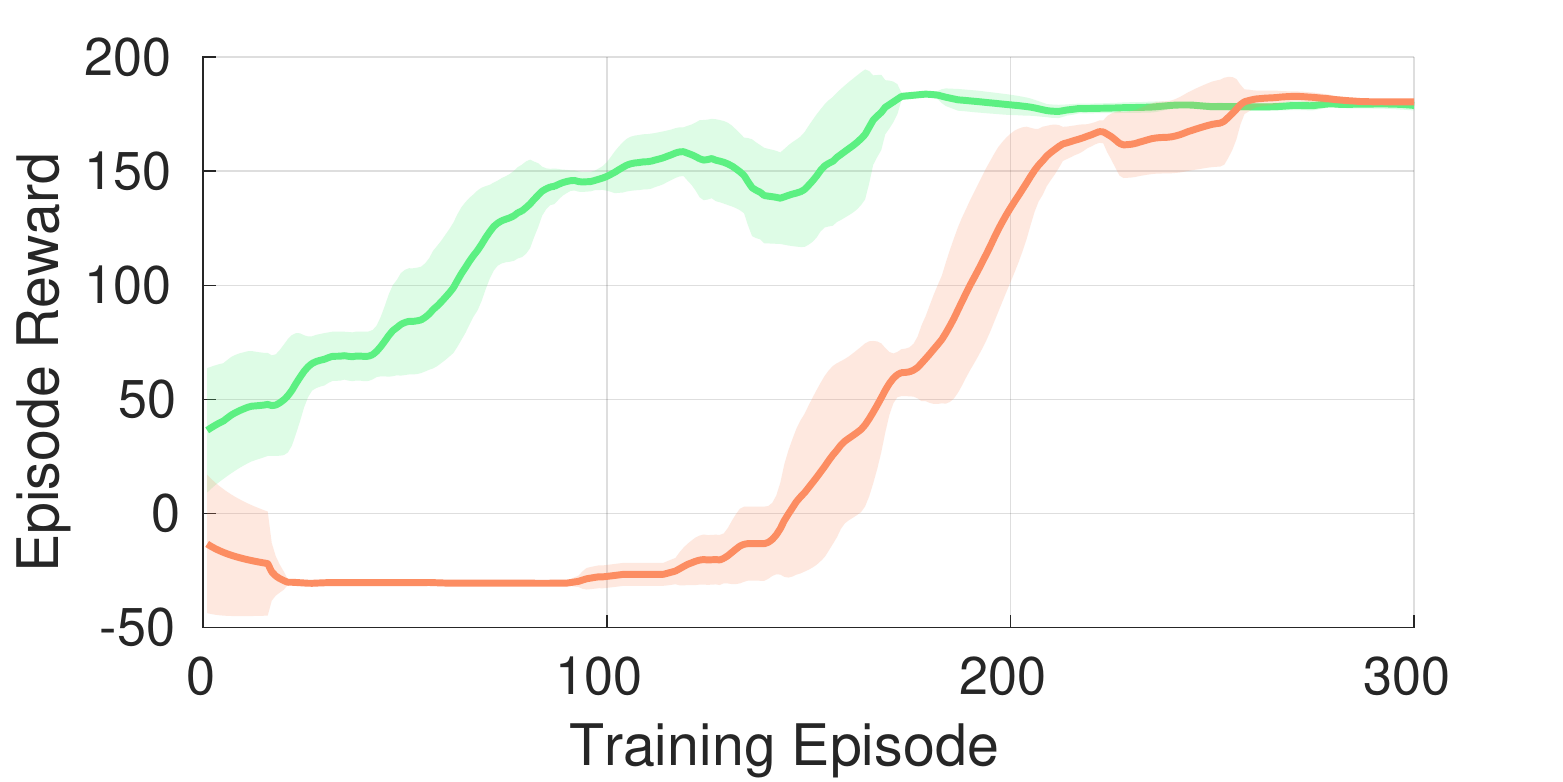}
  \caption{Running average of reward and its standard deviation for the cartpole swingup task.
  The RTS+RL framework (green) is able to more rapidly converge to a high reward policy when compared to a baseline RL approach (red).}
  \label{fig:train_cartpole}
\end{figure}

We summarize the results in Table \ref{tbl:cartpole}.
The reward during training for each RL agent is shown in Fig. \ref{fig:train_cartpole}.

The main result is that the RTS+RL agent achieves higher reward in fewer episodes.
Both agents are eventually able learn a policy to complete the swing up task, and both obtain similar reward on successful trials. 
However, only RTS+RL is able to guarantee safety while completing the task.
Surprisingly, RTD is unable to complete the task at all.
This is likely due to the non-convex reward; RTD becomes stuck in a local minimum because it is unable to optimize for long-term reward without a high-level planner.
In this local minimum, RTD keeps the cart moving back-and-forth and inadvertently maximizes the middle-of-the-track reward, avoiding the large possible minimum reward seen by RTS+RL and Baseline RL, which sometimes become stuck when trying to complete the global task.
This shows the importance of the planning hierarchy typically used for RTD \cite{kousik2018bridging}.
Note, RTS+RL's time per step (i.e. the time to run the while loop in Algorithm \ref{algo:saferl}) is less than $t\plan = 0.1$ s, so our method can be used to ensure RL safety online.

\begin{table*}[t]
\centering
\begin{tabular}{l|r|r|r}
\rowcolor[HTML]{C0C0C0} 
\multicolumn{1}{c|}{\cellcolor[HTML]{C0C0C0}{\textbf{Cartpole SwingUp Result}}} & \multicolumn{1}{c|}{\cellcolor[HTML]{C0C0C0}{\textbf{RTS+RL}}} & \multicolumn{1}{c|}{\cellcolor[HTML]{C0C0C0}\textbf{Baseline RL}} & \multicolumn{1}{c}{\cellcolor[HTML]{C0C0C0}\textbf{Reward RTD}} \\ \hline
\rowcolor[HTML]{FFFFFF} 
Avg. Planning Time {[}s{]}                                                                           & 0.027                                                                               & \textbf{8.6E-6}                                                   & 0.055                                                              \\ \hline
\rowcolor[HTML]{FFFFFF} 
Goals Reached {[}\%{]}                                                                               & \textbf{99.6}                                                                       & 98.2                                                              & 0.0                                                                \\
\rowcolor[HTML]{FFFFFF} 
Safely Stopped {[}\%{]}                                                                              & \textbf{0.4}                                                                        & \textbf{0.4}                                                      & 100.0                                                              \\
\rowcolor[HTML]{FFFFFF} 
Collisions {[}\%{]}                                                                                  & \textbf{0.0}                                                                        & 1.4                                                               & \textbf{0.0}                                                       \\
\rowcolor[HTML]{FFFFFF} 
Safety Interventions {[}\%{]}                                                                        & 0.08                                                                                & N/A                                                               & N/A                                                                \\ \hline
\rowcolor[HTML]{FFFFFF} 
Min/Mean/Max Reward                                                                                         & $-38.2/\mathbf{82.7}/\mathbf{100.9}$                                                                      & $-20.2/82.6/97.8$                                                             &$\mathbf{-3.1}/4.5/21.6$                                                              
\end{tabular}
\caption{Performance of RTS+RL versus baseline RL and RTD on the cartpole swing up task.}
\vspace*{-0.4cm}
\label{tbl:cartpole}
\end{table*}

\subsection{Car Implementation Details}

\subsubsection{High-Fidelity Model}
We use the following high-fidelity model, adapted from \cite{rasekhipour2016potential} to represent a car driving on a multi-lane road.
The state space of the vehicle is $x\hi = [p\lng,p\lat,\psi,v,\delta] \in X\hi$, with dynamics
\begin{align}\label{eq:car_high-fid_model}
\dot{x}\hi =
    \left[\begin{array}{c}
        v \cos\psi - v\lat \cos \psi \\
        v \sin\psi + v\lat \cos \psi \\
        \omega \\
        c_1 v + c_2 u_{1} \\
        c_3\cdot\left(u_{2}-\delta\right)
    \end{array}\right],
\end{align}
where
\begin{align}
    \omega &= \frac{v \tan \delta}{c_4 + c_5 v^{2}},\ \regtext{and}\\
    v\lat &= \omega\left(c_6 + c_7 v^{2}\right).
\end{align}
Here, $p\lng$, $p\lat$, and $\psi$ are the longitudinal, lateral position, and heading in the global reference frame; $v$ is the longitudinal velocity in the robot's local (body-fixed) coordinate frame ($v\lat$ is its lateral velocity), and $\delta$ is its steering angle.
The control inputs are $u = [u_1,u_2]\trans \in U$ are the acceleration command $u_1$ and steering torque $u_2$.
The values $c_1,\cdots,c_7 \in \R$ are constant model parameters.

We specify $\psi \in [-0.3,+0.3]$ rad, $v \in [0, 5]$ m/s, $\delta \in [-0.1,0.1]$ rad.
The control inputs are drawn from $u_1 \in [-4,4]$ m/s\ts{2} and $u_2 \in [-2,2]$ rad/s.

\subsubsection{Planning Model}
\begin{figure}[ht]
\centering
\includegraphics[trim={1cm 0 1cm 0},clip,width=\linewidth]{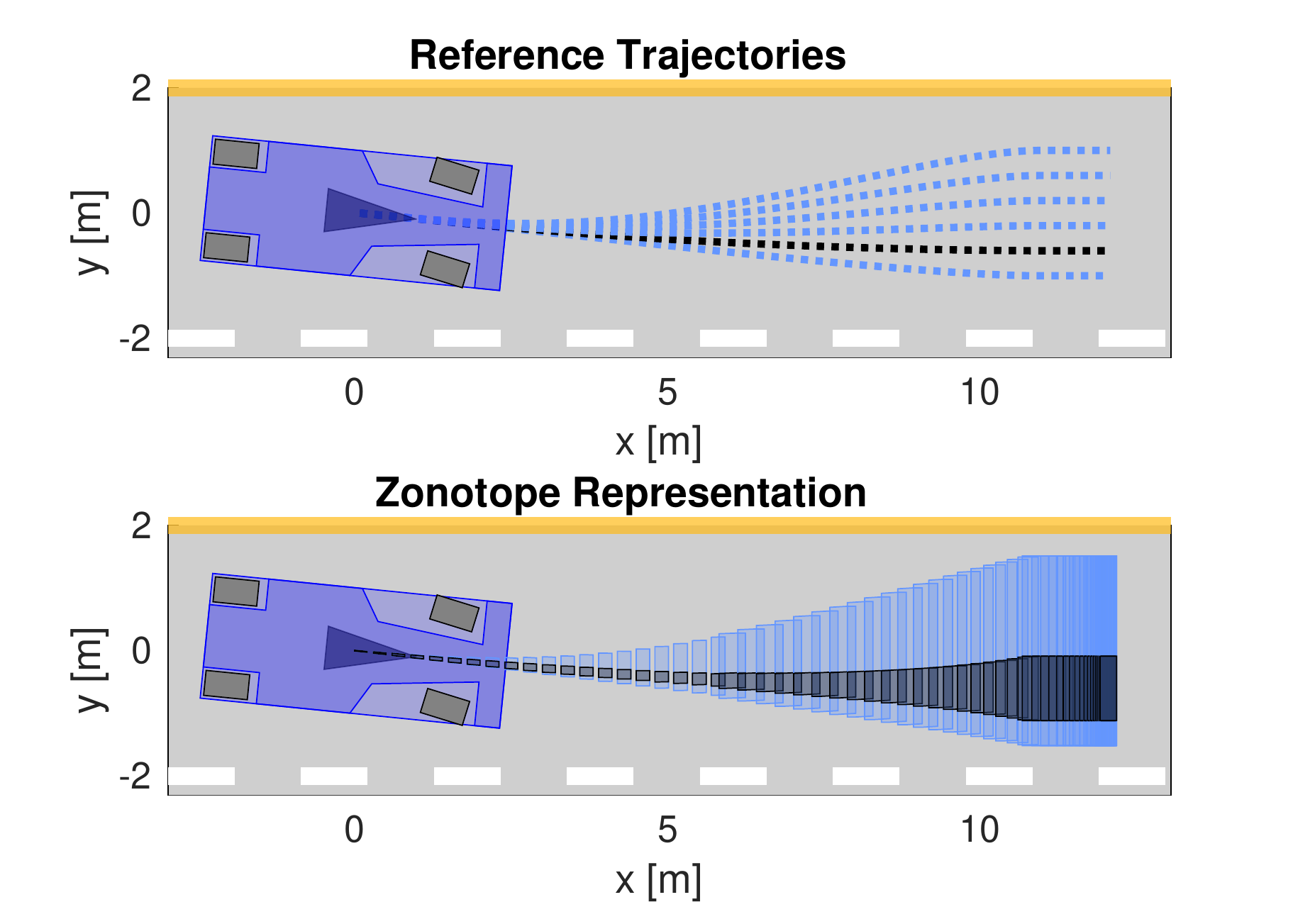}
\caption{The top subfigure shows plans for the car at different $k\desitwo$ parameter values ranging from $[-1,1]$ m, with one specific trajectory shown in black.
The bottom subfigure shows the zonotope PRS for the same range of $k\desitwo$.
The black subset is the sliced PRS corresponding to the trajectory in the top subfigure.
Notice that it conservatively contains the given trajectory.}
\label{fig:PRS}
\end{figure}

We use piecewise polynomials adapted from \cite{mueller_drone_motion}.
Recall that $p = (p\lng, p\lat)$.
Let $K\ini, K\desi \subset \R^2$, and denote $k = (k\inione, k\initwo, k\desione, k\desitwo) \in K$.
The parameter $k\desione$ specifies a velocity to be reached at a time $t\desione \in (0,t\final)$, and $k\desitwo$ specifies a lateral position to be reached at a time $t\desitwo \in (0,t\final)$.
Denote $p\plan = (p_1, p_2)$.
The planning model is given by
\begin{align}
    p_1(t,k) &= \frac{\tau_1(t, k)}{24} t^{4} + \frac{\tau_2(t, k)}{6} t^{3} + k\inione t \label{eq:v_lng_ref}, \\
    p_2(t,k) &= \frac{\tau_4(t, k)}{120} t^{5} + \frac{\tau_5(t, k)}{24} t^{4} + 
        \frac{\tau_6(t, k)}{6} t^{3} - \Delta_{v\lat}(k) t,\label{eq:v_lat_ref}
\end{align}
where $\tau_1,\cdots,\tau_7$ are given by
\begin{align}
    \begin{bmatrix}
        \tau_1(t,k) \\ \tau_2(t,k)
    \end{bmatrix} &=
    \frac{\Delta_{v\lng}(t,k)}{\left(\tau_3(t)\right)^{3}}
    \begin{bmatrix}
        -12  \\ 6 \tau_3(t)
    \end{bmatrix}\ \regtext{and}\\
    \begin{bmatrix}
        \tau_4(t,k) \\ \tau_5(t, k) \\ \tau_6(t,k)
    \end{bmatrix} &= \frac{1}{({\tau_7(t)})^{5}} 
        \begin{bmatrix}720 & -360 \tau_7(t)  \\
            -360 \tau_7(t) & 168 {\tau_7(t)}^{2}  \\
            60 {\tau_7(t)}^{2} & -24 {\tau_7(t)}^{3} 
        \end{bmatrix}
        \Delta\lat(t,k),
\end{align}
which are piecewise constant in $t$ because
\begin{align}
    \tau_3(t) &= \begin{cases}
            t\desione & t \in [0,t\desione) \\
            t\final - t\desione & t \in [t\desione,t\final]
        \end{cases},\\
    \tau_7(t) &= \begin{cases}
                t\desitwo & t\in [0,t\desitwo] \\
                t\final - t\desitwo & t\in [t\desitwo,t\final]
            \end{cases},
\end{align}
and each $\Delta_{(\cdot)}$ is given by
\begin{align}
    \Delta\lat(t,k) &= \begin{bmatrix}
            \Delta_{p\lat}(t,k) \\
            \Delta_{v\lat}(t,k) 
        \end{bmatrix},\\
    \Delta_{v\lng}(t, k) &= \begin{cases}
            k\desione - k\inione & t \in [0, t\desione) \\
            -k\desione           & t \in [t\desione, t\final]
        \end{cases},\\
    \Delta_{p\lat}(t, k) &= \begin{cases}
            k\desitwo - \Delta_{v\lat}(t,k) & t \in [0, t\desitwo) \\
          0          & t \in [t\desitwo, t\final]
            \end{cases},\ \regtext{and}\\
    \Delta_{v\lat}(t, k) &= \begin{cases}
            -k\inione\sin(k\initwo) & t \in [0, t\desitwo) \\
             0                      & t \in [t\desitwo, t\final]
            \end{cases}.
\end{align}
Example plans are shown in Figure \ref{fig:PRS}.

We specify $k\inione,k\desione \in [0,5]$ m/s and $k\initwo,k\desitwo \in [-1,1]$ m.
We set the timing parameters as $t\plan = 2$ s, $t\desione = 2$ s, $t\desitwo = 4$ s, and $t\final = 6$ s.
Note, as we see in the results in Sec. \ref{sec:experiments} of the paper, the car consistently finds new plans in $\approx 0.06$ s.
However, we chose $t\plan = 2$ s because we found that this reduced oscillations due to the RL agents choosing frequent lane changes, which improved the agents' ability to complete the car lane change task.

\subsubsection{Tracking Controller}
Recall that $P = P\lng\times P\lat$ and $x\hi = (p,\psi,v,\delta) \in \R^5$.
Let $p\plan$ be a plan with $k\ini$ determined by $x\hi$.
We specify $u\track$ for the car as a PD controller:
\begin{align}
    u\track(t,x\hi(t)) = \Gamma\cdot\left(\begin{bmatrix} p(t) \\ \psi(t) \\ v(t) \\ \delta(t)  \end{bmatrix} -
    \begin{bmatrix} p\plan(t,k) + p_0 \\ 0 \\ \dot{p}_1(t,k) \\ 0 \end{bmatrix}\right),
\end{align}
where $\Gamma \in \R^{2\times 5}$ is a matrix of control gains, $p_0 = \proj_P(x\hi(0))$, and $\dot{p}_1$ is the time derivative of \eqref{eq:v_lng_ref}.

\subsubsection{Reachability Hyperparameters}
To compute the PRS and ERS, we partition $T$ into $\numT = 120$ intervals of duration $\Delta_T = 0.05$ s.
We partition $k\inione$ into 2 intervals, and do not partition $k\initwo$, so $\numK = 2$.
We partition the $\theta$ dimension of $X\hiz$ into 13 intervals, the $v$ dimension into 2 intervals, and the $\delta$ dimension into 5 intervals, so $\numXhi = 13\times 2\times 5$.

\subsubsection{Observations}
We specify six elements for the car's observations.
The first four are the relative distance from the car's center of mass to the closest obstacle and second closest obstacle in front of the car, $\Delta\lng\arridx{1}$, $\Delta\lat\arridx{1}$, $\Delta\lng\arridx{2}$ and $\Delta\lat\arridx{2}$.
The last two observations are the car's velocity and global lateral position, $v$ and $p\lat$.
So,
\begin{align}
    o = \left(\Delta\lng\arridx{1},\Delta\lat\arridx{1},\Delta\lng\arridx{2},\Delta\lat\arridx{2},v,p\lat\right).
\end{align}

\subsubsection{Reward}
Recall the robot's state is $x\hi = (p\lng,p\lat,\psi,v,\delta)$.
The reward encourages the car to get to the goal, and some auxiliary rewards are added to help learning.
The total reward for each step consists of a speed reward, a lane reward, and goal reward:
\begin{align}
    r(x\hi,o) &=  r_{\mathrm{speed}}(x\hi) + r_{\mathrm{lane}}(x\hi,o) + r_{\mathrm{goal}}(x\hi)
\end{align}
The speed reward is
\begin{align}    
    r_{\mathrm{speed}}(x\hi) &= \exp\left(\frac{-1}{v^2+1}\right) - 3.7.
\end{align}
The lane reward encourages the car to stay in one of three lanes in the road-like environment.
Let $y\obs = p\lat + \delta\lat\arridx{1}$, where $\delta\lat\arridx{1}$ is part of the observation $o$ discussed above.
Then,
\begin{align}
    r_{\mathrm{lane}}(x\hi,o) &= \begin{cases}
    \rho_1(y\obs) - 2 \left|(p\lat-2)\right|-4& p\lat\in\left[0, 2\right)\\
    \rho_2(p\lat) - 4 & p\lat\in\left[2, 10\right) \\
    \rho_3(y\obs) - 2 \left|(p\lat-10)\right|-4& p\lat\in\left[10, 12\right]\\
  \end{cases},
\end{align}
where
\begin{align}
    \rho_1(y\obs) &= 5\exp\left(\frac{-1}{(y\obs-2)^{2}+1}\right), \\
    \rho_2(p\lat) &= 5\exp\left(\frac{-1}{{p\lat}^2+1}\right),\ \regtext{and}\\
    \rho_3(y\obs) &= 5\exp\left(\frac{-1}{(y\obs-10)^{2}+1}\right).
\end{align}
Finally, the goal reward encourages maintaining a high average velocity and reaching the end of the road
\begin{align}
    r_{\mathrm{goal}}(x\hi) &= \begin{cases}
        0 & p\lng < 500 \\
        100 & p\lng \geq 500.\\
    \end{cases}
\end{align}
Notice we require the car to attempt to drive for 500 m in each episode.\label{app:car}

\subsection{Drone Implementation Details}\label{app:drone}

\subsubsection{High-fidelity Model}
We use the following high-fidelity model for the drone \cite{lee_se3}:
\begin{align}\label{eq:drone_high_fid_mode}
    \ddot{p} &= \tau R e_3 - mge_3 \\
    \dot{R} &= R\hat{\omega} \\
    \dot{\omega} &= J\inv(\mu - \omega\times J\omega),
\end{align}
with states for position $p \in \R^3$, velocity $\dot{p} \in \R^3$, attitude $R \in \SO(3)$, and angular velocity $\omega \in \R^3$.
The model has been modified from \cite{lee_se3} to use a north-west-up convention for the global and body frame coordinate axes; $e_3$ is the global up direction.

Its control inputs $u = (\tau,\mu) \in U \subset R^4$ are thrust $\tau > 0$ and body moment $\mu \in \R^3$.
We convert these control inputs (given by a tracking controller described below) to rotor speeds using \cite[(3)]{kousik2019safe}, then saturate the rotor speeds to enforce compactness of $U$.

We use model parameters for an AscTec Hummingbird drone\footnote{The code is available online at \url{www.github.com/roahmlab/RTD_quadrotor_DSCC_2019}} with mass $m = 0.547$ kg and moment of inertia tensor $J = 10^{-3}\cdot\diag{3.3,3.3,5.8}$ kgm/s\ts{2}.
We set $g = 9.81$ m/s\ts{2}.
The rotor speeds (and therefore the control inputs) are saturated to within $[1100,8600]$ rpm.
See \cite[Table I]{kousik2019safe} for more details.
This model does not include aerodynamic drag, so it is valid up to a maximum speed of $\norm{\dot{p}(t)}_2 = 5$ m/s \cite{kousik2019safe}.
Further bounds on the states and inputs are in \cite{kousik2019safe}.

\subsubsection{Planning Model}
The drone uses the same planning model as the cartpole separately in each of the three directions of $\R^3$.
We bound its velocity parameter to $[-5,5]$ m/s, and its initial acceleration to $[-10,10]$ m/s\ts{2}.
We pick $t\plan = 1$ s, $t\desi = 1.5$ s, and $t\final = 3$ s.
See \cite{kousik2019safe} and \cite{shreyas_dissertation} for more details about the planning model, PRS computation, and ERS computation.

\subsubsection{Tracking Controller}
We use the tracking controller developed in \cite{lee_se3}, which has stronger stability guarantees than the controller used in \cite{kousik2019safe}, which we found in practice enables the RL agent to choose more aggressive trajectories while maintaining safety.
Per the notation in \cite{lee_se3}, we set the control gains as $k_x = 2$, $k_v = 0.5$, $k_R = 1$, and $k_{\Omega} = 0.03$.

\subsubsection{Observations}
We provide the drone RL agents with observations containing $35$ elements.
The first three are the drone's velocity in the global frame.
The next three are the unit vector $e_{\regtext{goal}}$ from the drone's global position to the global goal.
The remaining elements are the distances to obstacles along 29 rays extending from the drone's position (described below).
If any of these distances are greater than 10 m, we set them to 10 m.
So, one can write an observation for the drone as the tuple
\begin{align}
    o = \left(\dot{p},e_{\regtext{goal}},\{d\obs\idxi\}_{i=1}^{29}\right),
\end{align}
where each $d\obs\idxi$ is a distance along one of the 29 rays.
The rays are created using spherical coordinates.
We sample seven evenly-spaced azimuth angles in $[-3/16\pi,3/16\pi]$ rad, in addition to $-\pi/2$ and $\pi/2$ rad azimuth angles for looking sideways. We also use five elevation angles, at $-\pi/2, -\pi/12$, $0$, $\pi/12$ and $\pi/2$ rad. Lastly we omit the repeating rays to create 29 observation directions.

\subsubsection{Reward}
We use a goal reward to encourage getting to the goal and some auxiliary rewards to help learning:
\begin{align}
    r(x\hi,o) = r_v(x\hi,o) + r\obs(o) + r_{\regtext{goal}}(x\hi,o).
\end{align}
The first term rewards traveling in the direction of the goal:
\begin{align}
    r_v(x\hi,o) = \tfrac{1}{2}e_{\regtext{goal}}\cdot \dot{p}-2,
\end{align}
where $\cdot$ indicates the dot product.

The second term penalizes being near obstacles:
\begin{align}
    r\obs(o) = \left(\tan\inv\left(\mu\obs\right)\right)^4-4,
\end{align}
where $\mu\obs$ is the mean of the smallest eight observation distances from the drone's global position to the nearby obstacles (which may be a boundary of the world).

The final term rewards reaching the goal (i.e., traveling through the entire tunnel):
\begin{align}
    r_{\regtext{goal}}(x\hi) = \begin{cases}
        0 & p\arridx{1} < 97.5 \\
        100 & p\arridx{1} \geq 97.5,\\
    \end{cases}
\end{align}
recalling that $p \in \R^3$ is the drone's global position.




\end{document}